\documentclass{article} 
\usepackage{style,times}

\usepackage{amsmath,amsfonts,bm}

\def\eqref#1{equation~\ref{#1}}

\def\1{\bm{1}}

\def\vx{{\bm{x}}}
\def\vy{{\bm{y}}}

\DeclareMathAlphabet{\mathsfit}{\encodingdefault}{\sfdefault}{m}{sl}
\SetMathAlphabet{\mathsfit}{bold}{\encodingdefault}{\sfdefault}{bx}{n}

\newcommand\Eq[1]{Eq.\,\ref{#1}}
\newcommand\Fig[1]{{\textcolor{blue}{Fig.~\ref{#1}}}}

\newcommand\Tab[1]{Tab.~\ref{#1}}
\newcommand\App[1]{App.~\ref{#1}}

\renewcommand{\vx}{u_{x}}
\renewcommand{\vy}{u_{y}}
\newcommand{\bx}{B_{x}}
\newcommand{\by}{B_{y}}
\usepackage{hyperref}

\usepackage{hyperref}
\usepackage{url}
\usepackage{graphicx}
\usepackage{amsmath}
\usepackage{amssymb}
\usepackage{booktabs} 
\usepackage{makecell} 
\usepackage{multirow}
\usepackage{tabularx} 
\usepackage{array}  
\usepackage{bm}   
\usepackage{wrapfig}
\usepackage{cleveref}

\title{PHASE: Multi-Regime Modeling of Incompressible Magnetohydrodynamics}

\author{
Radhika Achikanath Chirakkara$^{1}$\thanks{ Correspondence to: rchirakkara@cita.utoronto.ca} \hspace{0.05cm},
Rajdeep Haldar$^{2}$\thanks{rhaldar@purdue.edu}\hspace{0.1cm},
Zezheng Song$^{3}$\thanks{zsong001@terpmail.umd.edu}\hspace{0.1cm},
and Jiequn Han$^{4}$\thanks{jhan@flatironinstitute.org}
\\
$^{1}$Canadian Institute for Theoretical Astrophysics, University of Toronto, ON, Canada, M5S 3H8\\
$^{2}$Department of Statistics, Purdue University, IN, United States of America, 47907\\
$^{3}$University of Maryland, College Park, MD, United States of America, 20742\\
$^{4}$Flatiron Institute, New York, NY, United States of America, 10010
}

\usepackage{xcolor}

\begin{document}
\iclrfinalcopy

\maketitle
\lhead{Preprint, under review.}

\begin{abstract}
Magnetohydrodynamics (MHD) is central to plasma modeling in astrophysics, space science, fusion, and engineering, but resolving multiscale MHD dynamics is computationally expensive. Machine-learning surrogates enable fast inference by learning reusable solution operators, yet existing models require separate training for each physical regime, limiting generalization across varying parameter settings. We introduce \textbf{PHASE}, a \emph{PHysics-Adaptive Scalable operator with residual Error correction}, designed to model  incompressible MHD across varying physical parameters with a single model. PHASE combines transfer learning, regime-aware adaptation, physics-centered learning, and residual refinement to improve both physical fidelity and generalization across MHD regimes. Together, these improvements achieve state-of-the-art prediction accuracy on two-dimensional MHD turbulence by reducing relative \(L_2\) errors on physical fields by more than an order of magnitude compared to prior MHD neural-operator baselines. Moreover, PHASE generalizes successfully to unseen parameter values without retraining, demonstrating the cross-regime adaptability expected from operator learning. 
We evaluate PHASE beyond point-wise prediction errors using derived physical fields, spectral analysis, and distribution statistics, consistently observing improved physical fidelity. We further show that our framework can accurately simulate MHD instabilities by testing it on the Kelvin--Helmholtz instability, demonstrating the robustness of our method.
\end{abstract}

\section{Introduction}

\label{sec:introduction}
Magnetohydrodynamics (MHD) describes the coupled evolution of electrically conducting fluids and magnetic fields, forming the basis for studying plasmas in astrophysics, space science, fusion, and engineering \citep{Kulsrud2005}. Predicting MHD dynamics is especially challenging in turbulent and instability-driven regimes, where nonlinear interactions between fluids and magnetic fields generate multiscale and intermittent structures. Traditionally, these dynamics are simulated by numerically solving the governing partial differential equations (PDEs) using finite-element methods \citep{Pencil2021, Athena++2020, Fryxelletal2000}. For many MHD problems, accurately resolving the relevant spatial and temporal scales is computationally expensive, and exploring different initial conditions and physical parameters requires many independent simulations.

Machine-learning-based surrogates instead learn a reusable solution map from initial states and physical parameters to resulting trajectories, amortizing simulation cost over many problem instances and enabling rapid inference. Neural operators (NO) provide a natural framework for this task by learning mappings between function spaces rather than solutions to individual problem instances \citep{Li_Kovachki+2020}. Physics-informed neural operators (PINO) further incorporate the governing equations during training, improving physical consistency and reducing reliance on simulation data \citep{Li_Peng+2022}. Nevertheless, accurately learning strongly turbulent and magnetically coupled dynamics remains challenging.

Current machine-learning surrogates for  incompressible MHD primarily employ physics-informed Fourier neural operators (FNO) \citep{Rosofsky&Huerta2023}. These models perform well in relatively smooth-laminar regimes but lose accuracy as the kinetic Reynolds number, \(Re\), increases and sharper turbulent structures develop. Recent advances augment FNO with conditional diffusion to predict turbulent dynamics \citep{Kacmaz+2025}. Despite these improvements, these models are trained separately for each \(Re\), limiting their ability to generalize across physical regimes or predict solutions at previously unseen parameter values. Evaluation is typically limited to velocity and magnetic fields in decaying MHD turbulence, which may not reveal errors in small-scale vortices and magnetic current sheets. These features are more directly characterized by their spatial derivatives, through the vorticity and current-density fields. Furthermore, it remains uncertain whether existing surrogates can model qualitatively distinct, instability-driven MHD dynamics.

\begin{figure*}[t]
    \centering
    \includegraphics[width=0.85\textwidth]{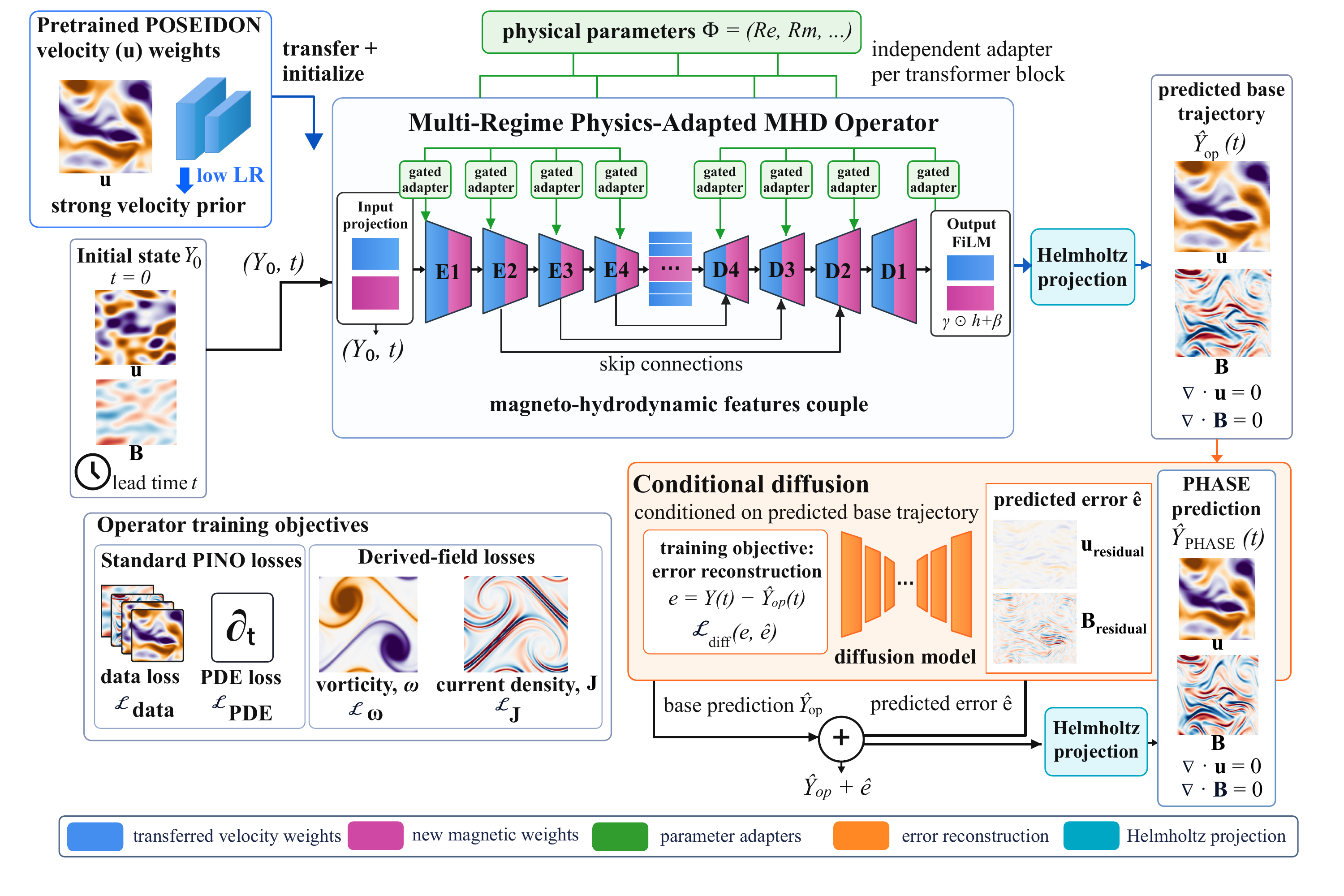}
    \caption{
    \textbf{Overview of PHASE.}
    PHASE initializes from pretrained fluid representations and augments them with MHD-specific magnetic field parameters. Regime-dependent adapters condition the operator on the physical parameters \((Re,Rm)\), while Helmholtz projection structurally enforces divergence-free constraints, and adding derived-field objectives improves physical fidelity. Finally, a conditional diffusion model learns the operator's residual error and refines unresolved structures. 
    }
    \label{fig:phase_overview}
\end{figure*}

Motivated by these limitations, we introduce \textbf{PHASE}, a \emph{PHysics-Adaptive Scalable operator with residual Error correction}. PHASE is formulated as a parameter-conditioned MHD operator whose internal representations adapt to changing physical regimes, rather than requiring a separate model for each parameter setting. We initialize the model using pretrained fluid representations from the scalable Operator Transformer (scOT) model in POSEIDON \citep{POSEIDON2024}, and extend the operator to learn coupled velocity--magnetic field dynamics. Physics-based structural constraints and objectives improve the consistency of the predicted fields and their small-scale structures, while a conditional diffusion model refines the residual errors left by the deterministic operator. The complete framework is presented in \autoref{sec:method} \& illustrated in \Fig{fig:phase_overview}.

PHASE is evaluated across both seen and unseen Reynolds numbers, as well as on freely decaying turbulence and instability-driven dynamics. In addition to point-wise field errors, the evaluation assesses whether the learned trajectories preserve physically meaningful derivative fields, multiscale spectra, distributional statistics, and divergence constraints, as described in \autoref{sec: experimental setup}. The results demonstrate that PHASE substantially outperforms SOTA FNO-based MHD surrogates, generalizes across physical regimes, and captures dynamics beyond the turbulence scenarios considered during model development \autoref{subsec: results}. The main contributions of this work are summarized as follows:
\begin{itemize}
\item \textbf{Transfer learning}: We show that pretrained fluid-dynamics representations can be effectively transferred to MHD by adapting POSEIDON's velocity prior and extending the scOT backbone to learn magnetic field evolution and velocity--magnetic field coupling. This cross-physics transfer yields \textbf{state-of-the-art performance on two-dimensional (2-D) incompressible MHD turbulence} while improving training efficiency (\autoref{sec:multiregime}, \ref{sec:turb_SOTA}).
\item \textbf{Multi-regime modeling}: We introduce parameter-conditioned residual adapters that \emph{adapt} the operator backbone to changing physical regimes, rather than relearning a separate model from scratch for each setting. This enables a \textbf{single physics-adapted model} to operate across laminar and strongly turbulent MHD regimes, improving prediction accuracy and \textbf{generalizing} correctly to unseen parameter values (\autoref{sec:multiregime}, \ref{sec:turb_SOTA}, \& \ref{sec:results_generalization}).

\item \textbf{Physics-centered learning}: We combine explicit loss supervision of derived fields such as vorticity \& current-density with hard divergence-free constraints through Helmholtz projection. Together, these improve the prediction of \textbf{small-scale structure, spectral fidelity, \& field distributions}, and enable accurate prediction beyond decaying turbulence to \textbf{instability-driven MHD dynamics} (\autoref{sec:physics_structure}, \ref{sec:results_instability}).

\end{itemize}

\section{Related Work}
\textbf{PDE foundation models \& transfer learning.} Recent work has explored foundation models for PDEs through multi-physics pretraining and transfer to downstream solution operators \citep{McCabe+2023, Hao+2024, Alkin+2024, POSEIDON2024, morel2025disco}. In particular, POSEIDON pretrains the multiscale scOT architecture on compressible and incompressible Navier--Stokes trajectories, achieving improved accuracy and sample efficiency across diverse PDE solutions \citep{POSEIDON2024}. However, the applicability of such foundation models to MHD remains unexplored.

\textbf{ML Surrogates for MHD.} \citet{Rosofsky&Huerta2023} introduced a PINO surrogate for 2-D incompressible MHD. Their tensorized Fourier neural operator (tFNO) combines training on MHD simulation data with losses derived from the governing equations and physical constraints, achieving correct predictions for laminar flows with $Re\leq250$. However, its performance degrades in turbulent regimes, where it struggles to learn small-scale velocity and magnetic field structures. Building on this work, \citet{Kacmaz+2025} combined the physics-informed tFNO with a conditional diffusion model, henceforth referred to as DINO. Although diffusion improves the recovery of turbulent structures, prediction errors still increase with $Re$, implying that strongly turbulent regimes remain challenging. Moreover, both frameworks train separate models for each Reynolds number and focus exclusively on decaying turbulence, leaving generalization across physical parameter regimes and applicability to other MHD problems largely unexplored.

\section{Background: Incompressible Magnetohydrodynamics}
\label{sec:mhd_background}
We consider the incompressible MHD equations, which describe the coupled evolution of a conducting fluid velocity field, \(\mathbf{u}\), and magnetic field, \(\mathbf{B}\):
\begin{align}
\partial_t \mathbf{u}
+ (\mathbf{u}\cdot\nabla)\mathbf{u}
&=
-\nabla\left(p+\frac{|\mathbf{B}|^2}{2}\right)
+(\mathbf{B}\cdot\nabla)\mathbf{B}
+\nu\nabla^2\mathbf{u},
\label{eq:momentum}
\\
\partial_t \mathbf{B}
+(\mathbf{u}\cdot\nabla)\mathbf{B}
&=
(\mathbf{B}\cdot\nabla)\mathbf{u}
+\eta\nabla^2\mathbf{B},
\label{eq:induction}
\\
\nabla\cdot\mathbf{u} &= 0,
\qquad
\nabla\cdot\mathbf{B} = 0.
\label{eq:divergence_constraints}
\end{align}
Here \(p\) is the fluid pressure, \(\nu\) is the kinematic viscosity, and \(\eta\) is the magnetic diffusivity. The first equation governs fluid motion, including feedback from the magnetic field, while the second governs how the magnetic field is advected and diffused by the flow. The divergence constraints enforce incompressible fluid motion and ensure no magnetic monopoles exist. The kinetic and magnetic Reynolds numbers,
\begin{equation}
    Re=\frac{UL}{\nu},
    \qquad
    Rm=\frac{UL}{\eta},
\end{equation}
control the balance between advection and dissipation in the velocity and magnetic fields, respectively. Here, \(U\) and \(L\) denote characteristic velocity and length scales, respectively. Larger \(Re\) and \(Rm\) correspond to weaker viscous and resistive dissipation, allowing finer spatial structures and stronger turbulent dynamics to develop. Smaller \(Re\) and \(Rm\) correspond to stronger dissipation regimes, which suppresses small-scale structures and produces laminar velocity fields and smooth magnetic fields. Their ratio defines the magnetic Prandtl number, $Pm = Rm/Re$. For our scope, we set \(Re=Rm\), and hence \(Pm=1\) in all our experiments.
Two derived fields are particularly important for characterizing small-scale MHD structure:
\begin{equation}
    \mathbf{\omega} = \nabla\times\mathbf{u},
    \qquad
    \mathbf{J} = \nabla\times\mathbf{B} .
\end{equation}
The vorticity, \(\omega\), measures local rotation of the fluid, while the current-density, \(\mathbf{J}\), measures spatial variation of the magnetic field. As both depend on spatial derivatives, they amplify small-scale prediction errors and therefore provide a stricter test of physical fidelity than the primary velocity and magnetic fields alone. 

\section{PHASE: Physics-Adapted MHD Operator}
\label{sec:method}
PHASE models the evolution of the 2-D incompressible MHD state,
$$
    Y(t) = \left(\mathbf{u}(t), \mathbf{B}(t)\right)
$$
as a parametrized solution operator $\mathcal{G}_{\theta}$ with initial state \(Y_0\), lead time \(t\), \& external physical parameters \(\boldsymbol{\Phi}\) as the input:
\begin{equation}
    \widehat{Y}(t)
    =
    \mathcal{G}_{\theta}
    \left(
    \mathbf{Y}_0,t,\boldsymbol{\Phi}
    \right),
    \label{eq:phase_operator}
\end{equation}
In the 2-D incompressible MHD context discussed in \autoref{sec:mhd_background},
\(\mathbf{u}=(u_x,u_y)\), \(\mathbf{B}=(B_x,B_y)\), and
\(\boldsymbol{\Phi}=(Re,Rm)\).  While we focus on 2-D problems in this work, we note that the PHASE framework can be readily extended to 3-D MHD.
The framework consists of a multi-regime physics-adapted operator transformer that predicts the primary MHD evolution, followed by a conditional diffusion model that corrects the remaining prediction error. PHASE additionally incorporates parameter conditioning, hard physical constraints and losses on derived fields, so that a single model can accurately represent multiple MHD regimes and problems. An overview is shown in \Fig{fig:phase_overview}. Next, we discuss each component of our framework.

\subsection{Transfer Learning and Multi-Regime Conditioning}
\label{sec:multiregime}
We leverage transfer learning (TL) from POSEIDON, whose scalable Operator Transformer (scOT) is pretrained on fluid-dynamics data \citep{POSEIDON2024}. PHASE initializes its velocity-related weights from the corresponding pretrained POSEIDON weights, providing a prior for fluid evolution, while newly initialized magnetic field weights extend the operator to jointly predict \((\mathbf{u},\mathbf{B})\). During MHD fine-tuning, we use a smaller learning rate for the transferred velocity weights and a larger learning rate for the magnetic field weights. This allows the pretrained fluid representations to adapt gradually to the MHD domain while prioritizing the learning of magnetic field evolution and velocity--magnetic field coupling. Further implementation and training details are provided in \App{app: transfer_learning}.

TL alone does not address variation across physical regimes. We therefore condition the operator on $\mathbf{\Phi}=(Re,Rm)$, allowing a single model to adapt its internal representations as the governing physical parameters change.
We introduce lightweight parameter-dependent residual adapters throughout the scOT hierarchy. For a hidden representation \(h_\ell\), the conditioned update takes the form
\begin{equation}
h_\ell
\leftarrow
h_\ell
+
g_\ell(\mathbf{\Phi})\odot E_\ell(h_\ell),
\label{eq:conditioned_adapter}
\end{equation}
where \(E_\ell\) is a bottleneck residual adapter and \(g_\ell\) is a learned gate determined by the physical parameters. 
A final feature-wise modulation adjusts the physical output channels: 
\begin{equation}
    \widehat{Y}
    =
    \widehat{Y}
    +
    \alpha_{\rm out}[\gamma(\mathbf{\Phi})\odot\widehat{Y}
    +
    \beta(\mathbf{\Phi})]
,
    \label{eq:output_film}
\end{equation}
where \(\gamma(\Phi)\) and \(\beta(\Phi)\) are learned shallow MLPs.  Together, the deep residual adapters and output modulation
allow a single operator to adapt its internal and output representations across
physical regimes. 
Eq. \ref{eq:conditioned_adapter} is inspired by the parameter-efficient adapters in ML literature, which introduce lightweight residual updates around a shared pretrained backbone \citep{pmlr-v97-houlsby19a}. We further condition these residual updates on the physical parameters through learned gates. Eq. \ref{eq:output_film} follows the FiLM principle of feature-wise affine conditioning \citep{perez2018film}, which allows external parameters to rescale and shift representation. See \App{subsec: param_conditioning} for conditioning details. This design preserves the pretrained operator while learning parameter-dependent corrections across laminar and turbulent regimes. Full adapter and optimization details are given in \App{app:training_details}. Moreover, we would like to highlight that naive external parameter conditioning by introducing additional input channels in the operator is less effective. 
The isolated effect of TL, our multi-regime conditioning \& other ablations on our results are depicted in \Fig{fig:model_ablation}.

\subsection{Physics-informed improvements}
\label{sec:physics_structure}
A 2-D magnetic field can be represented using an out-of-plane magnetic vector potential \(A\) as, $\mathbf{B} = \nabla\times(A\hat{\mathbf z})$.
This representation guarantees \(\nabla\cdot\mathbf{B}=0\) by construction and is used by MHD numerical solvers and prior ML surrogates \citep{Rosofsky&Huerta2023}. 
PHASE directly predicts the physical vector fields ($\mathbf{u}$, $\mathbf{B}$). This avoids reconstructing the magnetic field from a learned  potential, for which derived quantities such as the current-density require higher-order differentiation. This representation particularly  improves predictions of the magnetic field and the derived current-density field. In incompressible MHD, both velocity and magnetic fields must satisfy
\(\nabla\cdot\mathbf{u}=0\) and \(\nabla\cdot\mathbf{B}=0\). Following recent work on divergence-free projection \citep{Li+2026_Helmholtzprojection}, PHASE applies a Helmholtz projection to both predicted vector fields. For either \(\mathbf q\in{\mathbf u,\mathbf B}\), each Fourier mode is projected onto its divergence-free component:
\begin{equation}
\widehat{\mathbf q}_{\perp}(\mathbf k)
=
\left(
\mathbf I-\frac{\mathbf k\mathbf k^\top}{|\mathbf k|_2^2}
\right)
\widehat{\mathbf q}(\mathbf k).
\label{eq:helmholtz_projection}
\end{equation}

The projection enforces the divergence constraints up to high numerical precision and is applied to both the operator prediction and the diffusion-corrected output. We further train the deterministic operator with a physics-informed objective
\begin{equation}
\mathcal{L}_{\mathrm{PHASE}}
=
\lambda_{\mathrm{data}}\mathcal{L}_{\mathrm{data}}
+
\mathcal{L}_{\mathrm{ic}} + 
\lambda_{\mathrm{PDE}}\mathcal{L}_{\mathrm{PDE}}
+
\lambda_{\omega}\mathcal{L}_{\omega}
+
\lambda_{J}\mathcal{L}_{J},
\label{eq:phase_loss}
\end{equation}
where $\mathcal{L}_{\mathrm{PDE}}$ penalizes residuals of the governing MHD equations, while  $\mathcal{L}_{\mathrm{data}}$ penalizes errors in the primary fields. $\mathcal{L}_{\mathrm{ic}}$ penalizes the component-wise primary field prediction error at the initial time, $t=0$. Both are standard components of physics-informed surrogate training. The newly introduced losses, $\mathcal{L}_{\omega}$ and $\mathcal{L}_{J}$, supervise vorticity and current-density learning, respectively. These terms explicitly target derivative-sensitive, small-scale vorticity and current-density structures that may not be adequately captured by primary field reconstruction errors alone. We choose \(\lambda_{\rm data}\) and \(\lambda_{\rm PDE}\) based on working configurations in prior work, while tuning the new hyperparameters \(\lambda_{\omega}\) and \(\lambda_J\) (see \App{app:loss_details} and \ref{app:vorticity_current_weights} for further details). The contributions of direct magnetic field prediction, Helmholtz projection, and the derived-field-based physics loss are shown in \Fig{fig:model_ablation}. 

\subsection{Residual Diffusion Correction}
\label{sec:diffusion_refinement}

The neural operator captures the dominant deterministic evolution, but its remaining errors become increasingly structured in strongly turbulent regimes, where fine-scale vortices and magnetic field structures are difficult to reproduce. Inspired by prior MHD surrogates, we use a diffusion model to recover small-scale features unresolved by the neural operator prediction \citep{Kacmaz+2025,lippe2023pde}.

However, unlike \citet{Kacmaz+2025}, whose conditional diffusion model reconstructs the full MHD trajectory, we train the diffusion model to predict only the residual between the neural-operator prediction and the DNS trajectory, focusing on error calibration.
We model the residual between the operator prediction and the ground-truth trajectory as $e$, and train a conditional diffusion model to predict $\hat{e}$ while conditioning on the operator trajectory, as shown in \Fig{fig:phase_overview} \citep{Ho+2020_diffusion, Song+2020_diffusion, Kacmaz+2025}.
The residual and final PHASE prediction are
\begin{equation}
e = Y-\widehat{Y}_{\mathrm{op}},
\qquad
\widehat{Y}_{\mathrm{PHASE}}
=
\Pi_{\mathrm{div}}
\left(\widehat{Y}_{\mathrm{op}}+\hat{e}\right),
\label{eq:phase_final}
\end{equation}
where \(\Pi_{\mathrm{div}}\) denotes the Helmholtz projection (Eq.~\ref{eq:helmholtz_projection}) applied separately to the velocity and magnetic fields. By learning only the correction, the diffusion stage focuses its capacity on structures unresolved by the deterministic operator rather than relearning the full MHD evolution. Details of the diffusion objective is provided in \App{app:loss_details}.

\section{Experiments \& Results}
\subsection{Experimental Setup}
\label{sec: experimental setup}
\paragraph{Training Data and physical regimes.}
We generate ground-truth trajectories by solving the 2-D incompressible MHD equations with periodic boundary conditions using the Dedalus spectral solver \citep{Dedalus2020}. The simulations evolve the velocity components ($\vx,\vy$) and magnetic vector potential ($A$), from which the divergence-free magnetic field is recovered as $\mathbf{B} = \nabla \times A$. PHASE is trained directly on the four primary field channels ($\vx, \vy, \bx, \by$). All simulations use $Re=Rm$, corresponding to magnetic Prandtl number $Pm=1$. Numerical solver settings, initial conditions, and data-generation details are provided in \App{sec:Dedalus}. We generate data for freely decaying MHD turbulence and Kelvin-Helmholtz instability over \[
\begin{aligned}
Re_{\rm train} \in \mathcal{S}_{Re}
= \{&80,200,400,650,1000,1500, 2050,2750,3600,4500\},
\end{aligned}
\]
spanning viscous to strong turbulence regimes. These values are selected to cover the corresponding range of dissipation scales; the selection procedure is described in \App{sec:multiRe_selection}. Each trajectory begins from independently sampled divergence-free velocity and magnetic fields and evolves without external forcing. Using these settings, we test whether an ML surrogate can reproduce multiscale MHD dynamics across a broad range of physical regimes.

\begin{wrapfigure}{r}{0.52\textwidth}
    \centering
    \includegraphics[width=\linewidth]{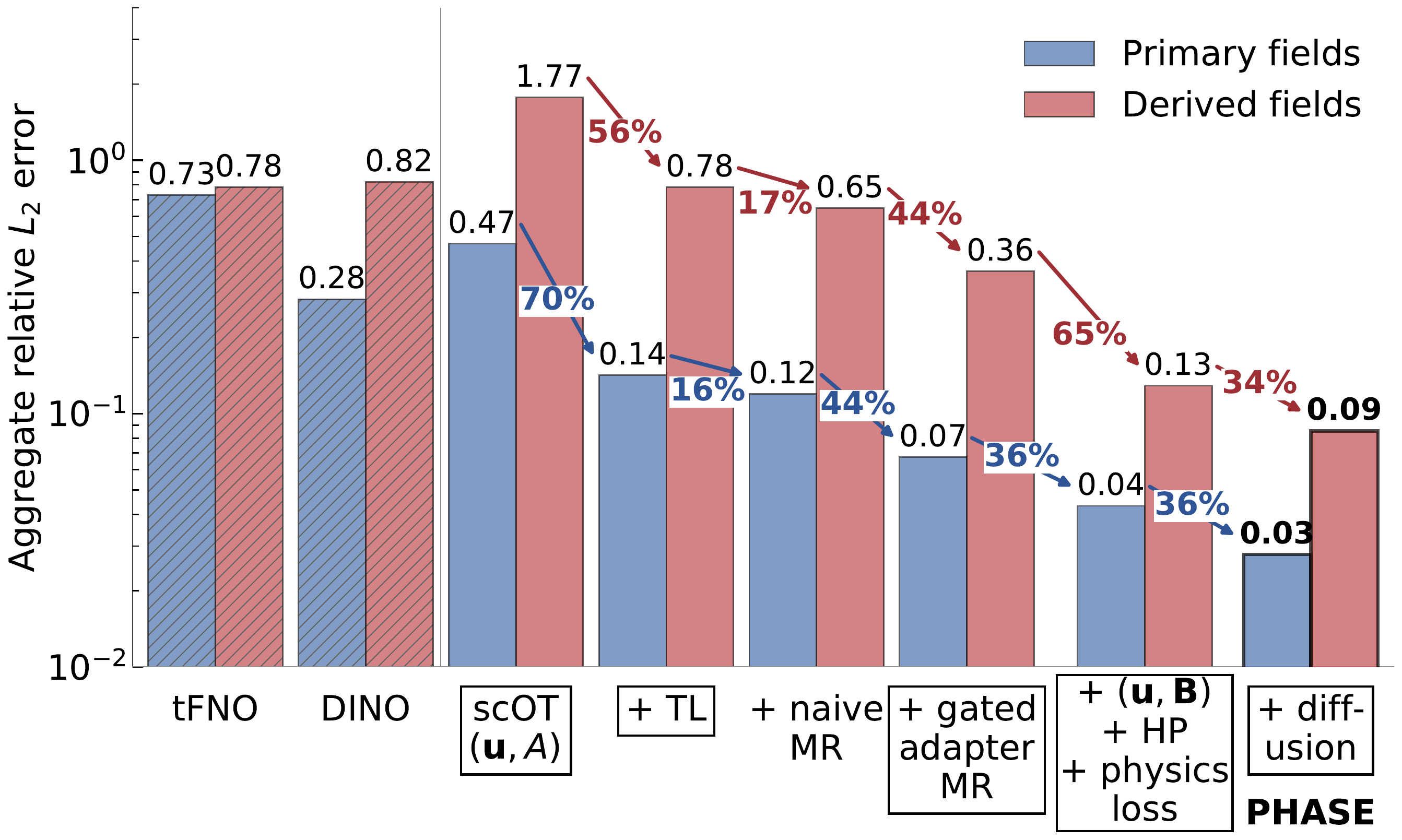}
    \caption{Aggregate relative \(L_2\) errors for the primary and derived fields across model ablations evaluated at \(Re=1000\), with percentage improvements between successive models. Hatched bars denote the previous tFNO and DINO baselines. TL, MR, and HP denote transfer learning, multi-regime training, and Helmholtz projection, respectively.}
    \label{fig:model_ablation}
\end{wrapfigure}

\paragraph{Baselines and training setup.} We compare PHASE against the physics-informed FNO surrogate of \citet{Rosofsky&Huerta2023} and the DINO model of \citet{Kacmaz+2025}, which augments a physics-informed tFNO with a conditional diffusion model. The latter DINO model represents the strongest prior baseline in our 2-D incompressible MHD setting. We perform an ablation study at $Re=1000$ to isolate the contributions of transfer learning, multi-regime conditioning, four-channel prediction, physics constraints, new derived-field-based physics loss and diffusion refinement. This is summarized in \Fig{fig:model_ablation}, which reports the aggregate relative $L_2$ errors for the primary fields \((u_x,u_y,B_x,B_y)\) and derived fields \((\omega,J)\) and the percentage improvement over the previously evaluated model (see \App{app:turb_ablation}). \Fig{fig:model_ablation} shows that the gains arise from the complete PHASE framework: the scOT backbone alone underperforms DINO, while transfer learning and gated multi-regime adapters provide successive improvements over  training from scratch and naive multi-regime conditioning. Four-channel prediction with Helmholtz projection and physics losses substantially improves the derived fields, and diffusion refinement produces the final PHASE model with the lowest primary- and derived-field errors. 
We evaluate both a single-regime (SR) PHASE trained on data for specific $Re$ and a multi-regime (MR) model trained jointly across the full Reynolds-number range $\mathcal{S}_{Re}$. The MR model is warm-started from the SR $Re=1000$ checkpoint before parameter-conditioned fine-tuning. Full architecture, optimization, and learning-rate details are provided in \App{app:training_details}.

\paragraph{Evaluation metrics.}
We evaluate both pointwise accuracy and physical fidelity. Relative \(L_2\) errors are reported for the primary velocity and magnetic fields and for the derived fields, vorticity and current-density, which provide a more sensitive measure of small-scale velocity and magnetic field structures. We also measure divergence error for both \(\mathbf{u}\) and \(\mathbf{B}\) (\App{app:turb_ablation}), multiscale spectral agreement, and distributional statistics including standard deviation and kurtosis (\App{app:turb_results_and_eval}). These complementary diagnostics test whether the learned MHD trajectories reproduce not only field values but also their spectral scaling, fluctuation amplitudes, and intermittent structures. 

\subsection{Results}
\label{subsec: results}
\subsubsection{State-of-the-Art Turbulence Prediction}
\label{sec:turb_SOTA}
\begin{figure}
    \centering
    \includegraphics[width=1.0\linewidth]{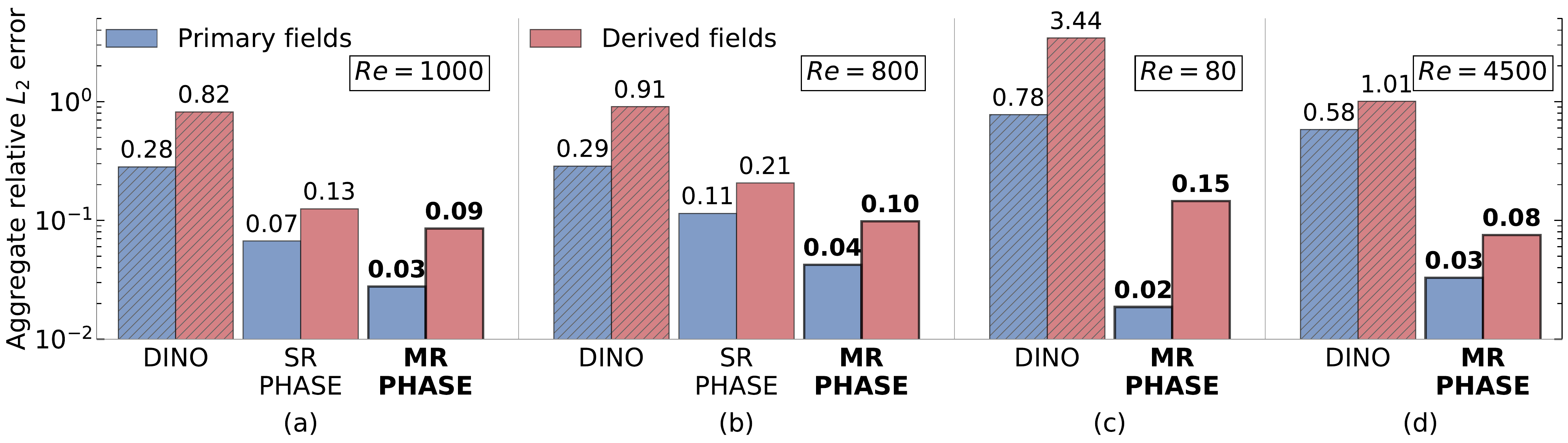}
    \caption{Aggregate relative \(L_2\) errors for the primary and derived fields. Panels (a)-(d) show results at \(Re=1000\), \(800\), \(80\), and \(4500\), respectively; the single-regime model is included for \(Re=1000\) and its unseen \(Re=800\) generalization case.}
    \label{fig:model_ablations_Re}
\end{figure}

First, we demonstrate that PHASE improves upon the previous state-of-the-art in predicting MHD dynamics in the turbulent regime.
\Fig{fig:model_ablations_Re} (a) shows the aggregate relative $L_2$ errors for the primary and derived fields in the turbulent regime with $Re=1000$, comparing the DINO model with SR and MR PHASE. MR PHASE performs best across all metrics, achieving roughly an order-of-magnitude reduction in errors across all fields compared to the previous state-of-the-art DINO model.

\begin{figure}
    \centering
    \includegraphics[width=0.72\linewidth,
        clip]{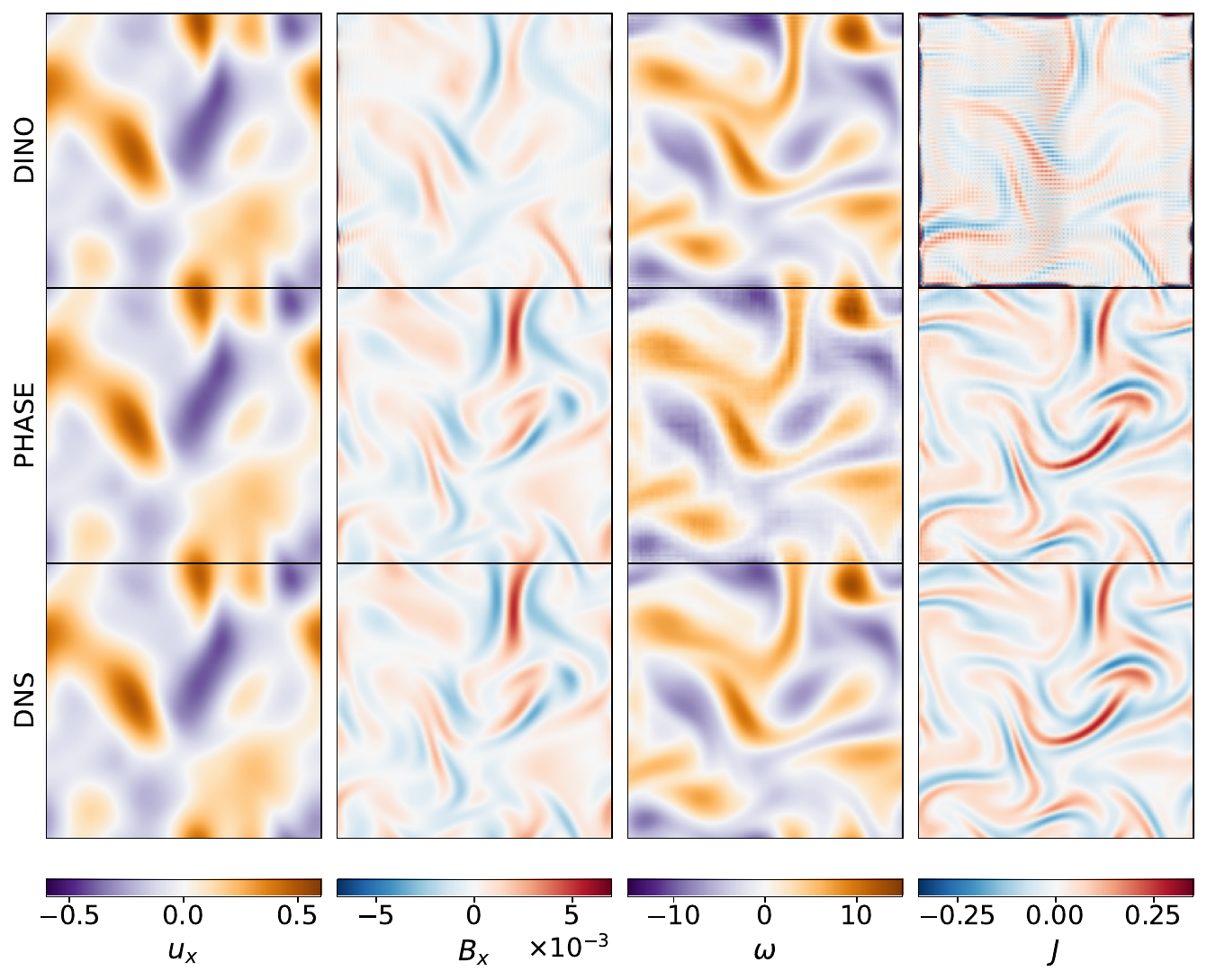}
    \caption{The $x$-components of the velocity and magnetic fields, vorticity, and current-density for $Re=1000$ at the final evolution time, $t=1.0$. The first row shows the DINO prediction, the second row shows the MR PHASE prediction, and the third row shows the DNS solution from the MHD simulation. We find that PHASE accurately captures structures in the primary and derived fields .}
    \label{fig:fields_Re1000}
    \vspace{-0.5 cm}
\end{figure}

\Fig{fig:fields_Re1000} compares the DINO and MR PHASE predictions with the DNS ground truth at the final simulation time, $t=1$. PHASE shows excellent agreement with the DNS for both primary and derived fields, whereas DINO exhibits pronounced inaccuracies, particularly in the current-density field. DINO predicts the magnetic vector potential, from which the magnetic field and current-density are obtained through first- and second-order spatial derivatives, respectively, amplifying errors. PHASE instead predicts both magnetic field components directly and includes dedicated physics losses for the derived fields (\autoref{sec:physics_structure}), capturing the magnetic field and current-density structures accurately. Although DINO ensures a divergence-free magnetic field through the vector-potential representation, it enforces velocity incompressibility through a soft loss, which does not guarantee divergence-free predictions. For the solutions shown, DINO yields $\nabla \cdot u \approx 0.15$. In contrast, Helmholtz projection (\autoref{sec:physics_structure}) yields $\nabla \cdot u \approx 9\times10^{-6}$ and $\nabla \cdot B \approx 6\times10^{-8}$ for PHASE, satisfying both constraints to high numerical accuracy.

\subsubsection*{Extremely viscous and turbulent regimes}

\begin{figure}
    \centering
    \includegraphics[width=0.9\linewidth,
        clip]{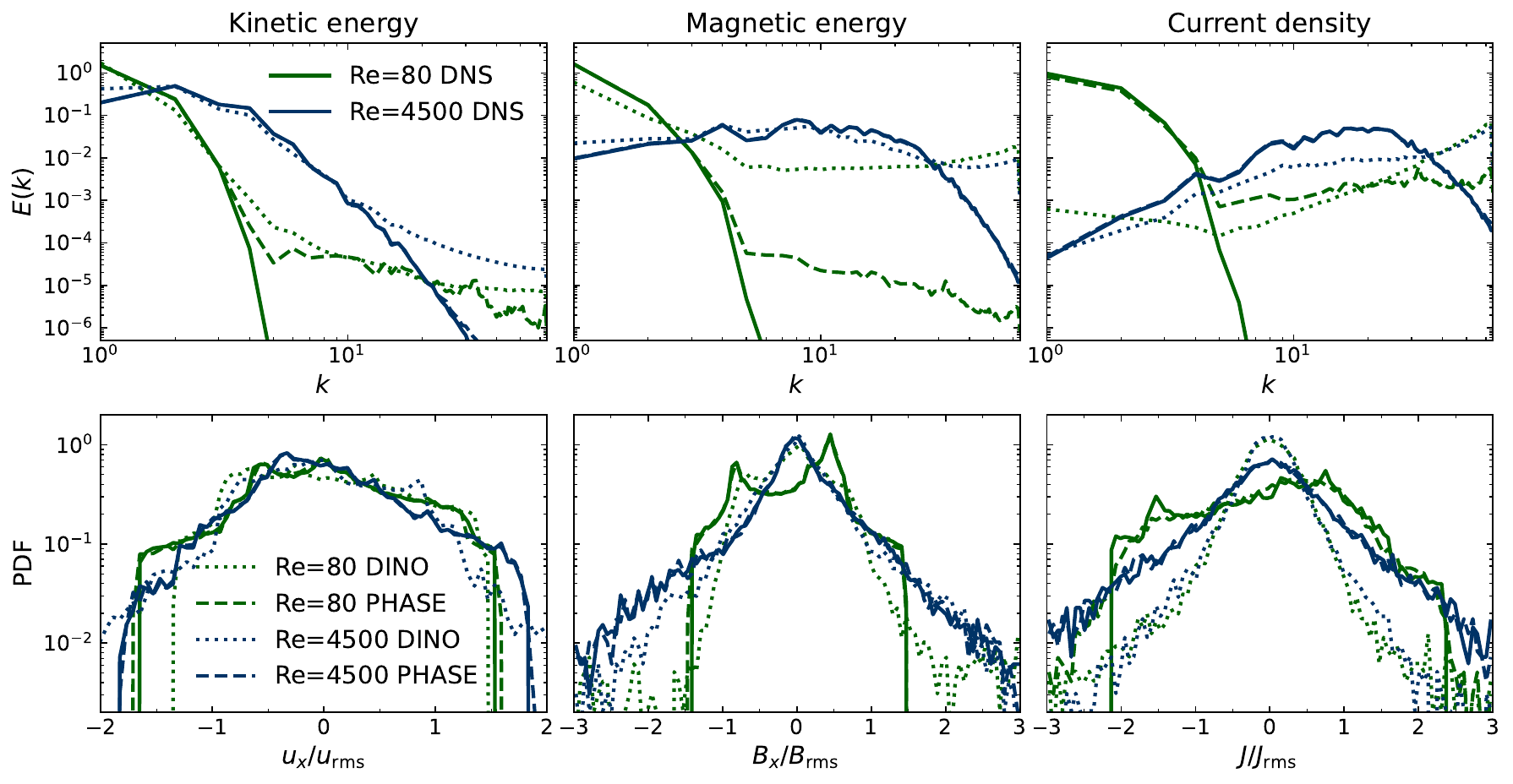}
    \caption{Spectra and PDFs for the DNS, DINO, and MR PHASE models at $Re=80$ and $4500$. The first row shows the normalized kinetic energy, magnetic energy, and current-density spectra, each scaled by its total spectral power. The second row shows the rms-normalized PDFs of $u_x$, $B_x$, and $J$. Compared to the DINO model, PHASE shows good predictions for all spectra and PDFs in the laminar and strongly turbulent regimes.}
    \label{fig:spectra_pdfs_Re80_Re4500}
\end{figure}

Next, we evaluate PHASE's performance in the  laminar and strongly turbulent regimes. In \Fig{fig:model_ablations_Re}, we show the DINO and MR PHASE performance in the viscous regime $(Re=Rm=80)$ and the strongly turbulent regime $(Re=Rm=4500)$. Our model achieves excellent performance in both extreme parameter regimes. To assess whether PHASE captures the distribution of kinetic and magnetic energy across spatial scales, we examine the kinetic energy, magnetic energy, and current-density spectra for $Re=80$ and $4500$ in \Fig{fig:spectra_pdfs_Re80_Re4500}, alongside the DINO predictions. We also examine the probability density functions (PDFs) of the primary and derived fields to evaluate if PHASE captures their statistical distributions and magnetic intermittency. The PDFs of $\vx$, $B_x$, and $J$ show excellent agreement with the DNS at both $Re=80$ and $Re=4500$. PHASE also reproduces the kinetic energy, magnetic energy, and current-density spectra well, particularly in the strongly turbulent $Re=4500$ regime, where the DINO model fails. At high wavenumbers (small length-scales), however, PHASE over-predicts the spectral power across all three spectra and this limitation is most pronounced at $Re=80$. Further discussion on the spectra and PDFs are provided in \App{sec:turb_spectra_pdfs}.

\subsubsection{Generalization Across Reynolds Number}
\label{sec:results_generalization}
The generalization of the $Re=1000$ trained DINO and SR PHASE models, together with MR PHASE, to an unseen Reynolds number, $Re=800$, is evaluated in \Fig{fig:model_ablations_Re} (b). While DINO fails to capture the dynamics at an unseen $Re$, both PHASE models perform substantially better. MR PHASE achieves the lowest errors across all primary and derived fields, demonstrating the strongest generalization to unseen Reynolds numbers (see \App{sec:turb_relL2} and \ref{sec:turb_spectra_pdfs}). This shows that unlike traditional MHD solvers, which require a separate simulation for each parameter value, PHASE can predict correct dynamics at unseen physical parameters correctly without retraining or running new numerical simulations. This generalization capability is a key advantage of our ML surrogate.

\subsubsection{Instability Dynamics}
\label{sec:results_instability}
\begin{figure}
    \centering
    \includegraphics[width=0.9\linewidth,
        clip]{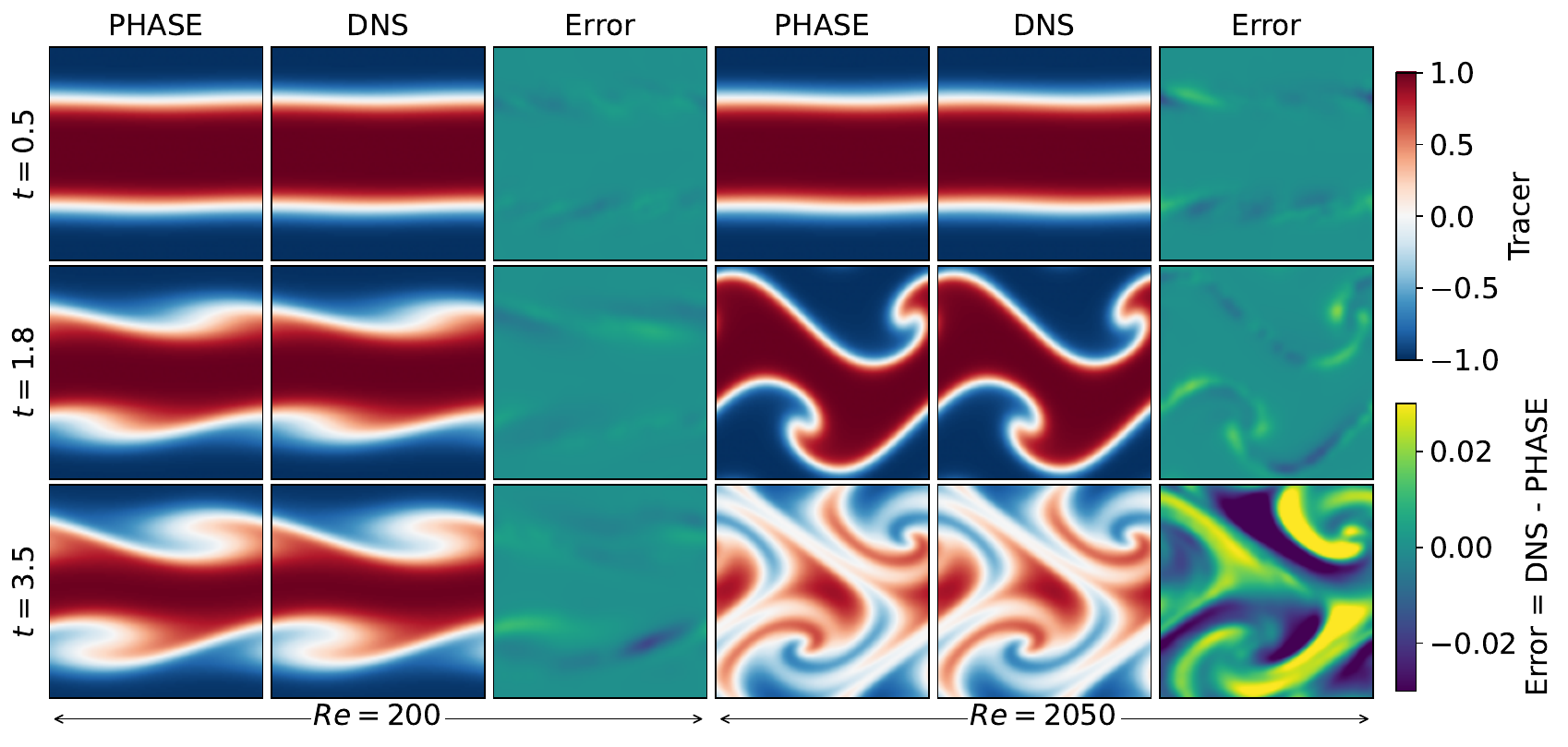}
    \caption{Tracer evolution for the KH instability test in the viscous ($Re= 200$) and turbulent ($Re=2050$) regime. Columns compare PHASE predictions, DNS, and errors \((\mathrm{DNS}-\mathrm{PHASE})\), while rows show the initial phase ($t=0.5$), linear growth ($t=1.8$), and the nonlinear mixing stage ($t=3.5$) of the KH instability. PHASE can accurately capture spatial and temporal KH instability dynamics across viscous and turbulent regimes.}
    \label{fig:KH}
\end{figure}

To test whether PHASE can model dynamics beyond freely decaying turbulence, we train and evaluate PHASE on the MHD Kelvin--Helmholtz (KH) instability. KH provides a more stringent test because the weak transverse perturbation velocity must be resolved in the presence of a much stronger shear flow, unlike in isotropic turbulence, where the two velocity components have comparable amplitudes. The model must capture the amplification of this perturbation through the linear KH growth stage, nonlinear vortex roll-up, saturation, and the subsequent decay of the instability. We evaluate PHASE in both viscous and turbulent KH regimes. To examine these different stages, we track a passive tracer that is advected and diffused by the flow; further details are provided in \App{sec:Dedalus}. The tracer provides a direct visualization of shear-layer deformation ($t=0.5$), vortex formation ($t=1.8$), and mixing ($t=3.5$). \Fig{fig:KH} compares the PHASE predictions, DNS solutions, and corresponding errors for the viscous ($Re=200$) and turbulent ($Re=2050$) regimes. PHASE closely reproduces the tracer evolution in both regimes, capturing the entire time evolution of the KH instability (see \App{app:kh_results} for further details).

\section{Conclusion and Future Work}
We introduced PHASE, a physics-adapted multi-regime neural operator for incompressible MHD. PHASE transfers pretrained velocity representations from the fluid-dynamics foundation model POSEIDON and extends the scOT backbone to learn magnetic field evolution. Together with residual diffusion refinement and physics-structured improvements, this model achieves state-of-the-art accuracy on MHD turbulence predictions, reducing relative $L_2$ errors by more than an order of magnitude compared with previous MHD neural-operator surrogates from \cite{Kacmaz+2025}. 

Parameter-conditioned residual adapters enable a single model to adapt across laminar and strongly turbulent MHD regimes and generalize to unseen parameter values without retraining. 
Multi-regime modeling not only improves cross-regime generalization but also outperforms the corresponding single-regime model at a fixed Reynolds number. Multi-regime training is therefore beneficial even when the model is applied only to a single physical regime.
Explicit losses on vorticity and current-density further improve the recovery of small-scale structures, spectral scaling, and field distributions. PHASE also captures the evolution of the Kelvin--Helmholtz instability, demonstrating applicability beyond decaying turbulence to instability-driven dynamics. These results establish PHASE as a step toward a unified MHD model across physical regimes. 

PHASE is currently developed and evaluated only for 2-D incompressible MHD. This study does not address compressibility, supersonic flows, shock formation, or three-dimensional dynamics. We also restrict our experiments to $Re=Rm$. Therefore, the model's performance when viscous and resistive scales vary independently remains unexplored. As found in \Fig{fig:spectra_pdfs_Re80_Re4500}, PHASE over-predicts spectral power at high wavenumbers. Future work will investigate targeted spectral losses and reinforcement-learning-based refinement to improve these predictions. We will also extend PHASE to compressible MHD, including supersonic Mach number regimes, varying magnetic Prandtl numbers, and broader classes of plasma instabilities.

\subsection*{Code \& Data availability}
We provide details pertaining to data generation, loss \& training details in \App{sec:Dedalus}, \App{app:loss_details} \& \App{app:training_details} respectively. Readers interested in the code implementation can refer to the public link to the PHASE repository on github \href{https://github.com/PHASE-MHD/PHASE}{[link]}. Furthermore, one can find the trained model weights for MR PHASE \href{https://huggingface.co/phaseMHD}{[here]}. 

\bibliography{refs}
\bibliographystyle{refs}
\newpage
\appendix
\section{Appendix}

\subsection{MHD Data generation}
\label{sec:Dedalus}
We solve the 2-D incompressible MHD equations on the periodic unit square, $(x,y)\in[0,1]^2$, using Dedalus \citep{Dedalus2020}. The spatial discretization uses $N_x=N_y=128$. At each time step, the solver evolves the velocity components $u_x$ and $u_y$ and the out-of-plane magnetic vector potential $A$. The in-plane magnetic field is reconstructed as $\mathbf{B}= \nabla\times A$, which satisfies $\nabla\cdot\mathbf{B}=0$ by construction. Although the simulations evolve $(u_x,u_y, A)$, the PHASE model uses the four physical field components $(u_x,u_y,B_x,B_y)$. For visualization, we additionally evolve a passive tracer, $s$, according to
\begin{equation}
\partial_t s+\mathbf{u}\cdot\nabla s=\nu\nabla^2s,
\end{equation}
which we particularly use to study instabilities. The tracer is not included among the input or target channels of PHASE.

For decaying turbulence, the initial conditions are sampled from smooth Gaussian random fields, following the prescription in \cite{Rosofsky&Huerta2023}. The fields are generated in Fourier space and satisfy periodic boundary conditions. We sample a stream function $\psi$ and magnetic potential $A$ and construct\begin{equation}\mathbf{u}=\nabla\times(\psi\hat{\mathbf{z}})=\left(\partial_y\psi,-\partial_x\psi\right),\qquad\mathbf{B}=\nabla\times(A\hat{\mathbf{z}})=\left(\partial_y A,-\partial_x A\right).\end{equation}This construction ensures that both initial fields satisfy their divergence-free constraints. For each physical-parameter configuration, we generate $1{,}000$ trajectories with independently sampled initial conditions. The code uses the fourth-order Runge--Kutta time stepping with $\Delta t=10^{-3}$. For all our simulations we use $Re=Rm$, i.e. $Pm = 1$

The decaying-turbulence trajectories are evolved without external forcing, so the flow evolves freely under nonlinear advection, Lorentz coupling, viscosity, and magnetic diffusion. Each trajectory is integrated until $t=1$, and solution snapshots are written every ten solver steps, giving an output timestep of $\Delta t_{\mathrm{out}}=10^{-2}$.

For the Kelvin--Helmholtz (KH) instability, we use the same periodic BCs, but replace the random velocity initialization by a two-layer shear profile. The streamwise velocity is initialized as
\begin{equation}
u_x(y)=U_0\left[
\tanh\left(\frac{y-y_1}{\delta}\right)
-\tanh\left(\frac{y-y_2}{\delta}\right)-1
\right],
\end{equation}
with shear layers centered at $y_1=1/4$ and $y_2=3/4$. The transverse velocity is seeded with a small perturbation to trigger the instability. In our runs we use $U_0=1$, perturbation amplitude $\epsilon=10^{-2}$, and shear-layer width $\delta=0.05$. The magnetic field is initialized as a divergence-free random field through the same magnetic potential construction used for the turbulence data. The KH trajectories are evolved until $t=5$, to capture the growth and saturation of the instability.

\subsection{Multi-regime Reynolds number selection}
\label{sec:multiRe_selection}
As the plasma is described by 2-D incompressible MHD, and the
magnetic field is extremely weak compared to the velocity field, the small-scale cutoff is estimated using the 2-D enstrophy-dissipation scale,
\begin{equation}
\chi
=
\nu\left\langle|\nabla\omega|^2\right\rangle_{x,y,t},
\end{equation}
where the average is taken over space and time. The corresponding viscous scale is
\begin{equation}
\ell_\chi=\left(\frac{\nu^3}{\chi}\right)^{1/6}.
\end{equation}
For our unit periodic domain, $k_{\mathrm{phys}}=2\pi k_{\mathrm{shell}}$;
therefore, the cutoff is
\begin{equation}
k_\chi
=
\frac{1}{2\pi}
\left(\frac{\chi}{\nu^3}\right)^{1/6}.
\end{equation}
We select
$Re\in\{80,200,400,650,1000,1500,2050,2750,3600,4500\}$
to approximately uniformly sample the estimated 
enstrophy-dissipation cutoff wavenumber $k_\chi$, spanning viscous to strongly
turbulent regimes. We restrict our study to $Pm=Rm/Re=1$, and therefore set
$Rm=Re$.
\subsection{Ablation table}
\label{app:turb_ablation}
We compare PHASE with the tFNO and DINO baselines from previous MHD surrogate studies \citep{Rosofsky&Huerta2023, Kacmaz+2025}. Unlike the training procedure in \citet{Rosofsky&Huerta2023}, we train tFNO from scratch, without a warm start, to enable a direct comparison with the scOT baseline trained without transfer learning or warm start. Detailed ablation results are reported in \Tab{tab:model_ablation_detailed}. The ablation study isolates the effects of velocity-weight transfer from POSEIDON; naive and gated-adapter multi-regime conditioning; the combined transition from three-channel $(u_x,u_y,A)$ to four-channel $(u_x,u_y,B_x,B_y)$ prediction with Helmholtz projection and physics losses on vorticity and current-density; and residual-error correction using diffusion.

We quantify prediction accuracy using relative $L_2$ errors. The component-averaged velocity and magnetic field errors are
$u_{\rm avg}=\tfrac{1}{2}[L_2(u_x)+L_2(u_y)]$ and
$B_{\rm avg}=\tfrac{1}{2}[L_2(B_x)+L_2(B_y)]$, respectively. The aggregate primary-field error is
$\mathbf{P}=\tfrac{1}{2}[u_{\rm avg}+B_{\rm avg}]$, while the aggregate derived-field error is
$\mathbf{D}=\tfrac{1}{2}[L_2(\omega)+L_2(J)]$. Within the ablation sequence, each successive addition improves both aggregate errors, as shown in \Fig{fig:model_ablation}. The final two columns of \Tab{tab:model_ablation_detailed} report the root-mean-square divergence errors for the velocity and magnetic fields. Helmholtz projection enforces both velocity incompressibility and the divergence-free magnetic field constraint to high numerical accuracy.

\begin{table*}[t]
\centering
\caption{Field-wise and aggregate relative $L_2$ errors for the model ablations. The tFNO and DINO results are the baselines from previous studies. We define $u_{\rm avg}=\tfrac{1}{2}[L_2(u_x)+L_2(u_y)]$ and
$B_{\rm avg}=\tfrac{1}{2}[L_2(B_x)+L_2(B_y)]$.
The primary field aggregate is
$\mathbf{P}=\tfrac{1}{2}[u_{\rm avg} + B_{\rm avg}]$,
and the derived aggregate is
$\mathbf{D}=\tfrac{1}{2}[L_2(\omega)+L_2(J)]$. The final two columns report
the root-mean-square divergence errors.}
\label{tab:model_ablation_detailed}
\setlength{\tabcolsep}{4pt}
\renewcommand{\arraystretch}{1.20}
\resizebox{\textwidth}{!}{%
\begin{tabular}{clcccccccc}
\toprule
\multirow{2}{*}{\textbf{Ser. No.}}
& \multirow{2}{*}{\textbf{Model}}
& \multicolumn{4}{c}{\textbf{Avg. rel. $L_2$}}
& \multicolumn{2}{c}{\textbf{Aggregate rel. $L_2$}}
& \multicolumn{2}{c}{\textbf{RMS div. errors}} \\
\cmidrule(lr){3-6}\cmidrule(lr){7-8}\cmidrule(lr){9-10}
&
& $u_{\rm avg}$
& $B_{\rm avg}$
& $\boldsymbol{\omega}$
& $\mathbf{J}$
& $\mathbf{P}$
& $\mathbf{D}$
& $\nabla\!\cdot\!\mathbf{u}$
& $\nabla\!\cdot\!\mathbf{B}$ \\
\midrule
1 & tFNO
& $0.523$ & $0.935$
& $0.571$ & $0.995$
& $0.729$ & $0.783$ & $2.3\times10^{-1}$ & $3.2\times10^{-8}$ \\
2 & DINO
& $0.035$ & $0.530$
& $0.081$ & $1.561$
& $0.283$ & $0.821$ & $1.5\times10^{-1}$ & $6.3\times10^{-8}$ \\
3 & scOT $(\mathbf{u},A)$
& $0.261$ & $0.676$
& $0.530$ & $3.009$
& $0.468$ & $1.770$ & $1.6\times10^{0}$ & $7.8\times10^{-8}$ \\
4 & 3 + TL
& $0.033$ & $0.250$
& $0.135$ & $1.430$
& $0.142$ & $0.782$ & $4.0\times10^{-1}$ & $7.6\times10^{-8}$ \\
5 & 4 + naive MR
& $0.028$ & $0.210$
& $0.126$ & $1.168$
& $0.119$ & $0.647$ & $3.8\times10^{-1}$ & $7.6\times10^{-8}$ \\
6 & 4 + gated adapter MR
& $0.020$ & $0.115$
& $0.098$ & $0.631$
& $0.067$ & $0.364$ & $3.2\times10^{-1}$ & $7.5\times10^{-8}$ \\
7 & \makecell[l]{6 + $(\mathbf{u},\mathbf{B})$ \\ + Helmholtz projection \\ + physics loss}
& $0.025$ & $0.062$
& $0.114$ & $0.143$
& $0.043$ & $0.129$ & $9.1\times10^{-6}$ & $6.1\times10^{-8}$ \\
8 & \textbf{PHASE}: 7 + diffusion
& $\mathbf{0.012}$ & $\mathbf{0.044}$
& $\mathbf{0.063}$ & $\mathbf{0.108}$
& $\mathbf{0.028}$ & $\mathbf{0.085}$
& $9.1\times10^{-6}$ & $6.0\times10^{-8}$ \\
\bottomrule
\end{tabular}%
}
\end{table*}

\subsection{Loss Details}
\label{app:loss_details}
\paragraph{Operator training objective.}
The backbone operator model predicts the four physical fields
\((u_x,u_y,B_x,B_y)\). Its total training loss is a weighted sum of data,
initial-condition, PDE-residual, vorticity, and current-density losses:
\begin{equation}
\mathcal{L}_{op}
=
10\,\mathcal{L}_{\mathrm{data}}
+ \mathcal{L}_{\mathrm{ic}}
+ 10^{-3}\,\mathcal{L}_{\mathrm{PDE}}
+ 2\,\mathcal{L}_{\omega}
+ 5\,\mathcal{L}_{J}.
\end{equation}
The vorticity and current-density loss weights are examined in \App{app:vorticity_current_weights}. The data loss is a weighted sum of component-wise relative \(L_2\) losses,
\begin{equation}
\mathcal{L}_{\mathrm{data}}
=
\ell(u_{x,\mathrm{pred}},u_{x,\mathrm{DNS}})
+ \ell(u_{y,\mathrm{pred}},u_{y,\mathrm{DNS}})
+ 5\,\ell(B_{x,\mathrm{pred}},B_{x,\mathrm{DNS}})
+ 5\,\ell(B_{y,\mathrm{pred}},B_{y,\mathrm{DNS}}),
\end{equation}
where
\begin{equation}
\ell(a_{\mathrm{pred}},a_{\mathrm{DNS}})
=
\frac{
\|a_{\mathrm{pred}}-a_{\mathrm{DNS}}\|_2
}{
\|a_{\mathrm{DNS}}\|_2
},
\end{equation}
where the direct numerical simulations (DNS) is the ground truth from the MHD simulations.
The initial-condition loss \(\mathcal{L}_{\mathrm{ic}}\) has the same component-wise form, but is evaluated only at the initial time. The vorticity and current-density losses are also relative \(L_2\) penalties.

The PDE loss penalizes the residuals of the four MHD evolution equations,
$(D_{u_x},D_{u_y},D_{B_x},D_{B_y})$, evaluated from the predicted fields:
\begin{equation}
\mathcal{L}_{\mathrm{PDE}}
=
\operatorname{MSE}(D_{u_x},0)
+
\operatorname{MSE}(D_{u_y},0)
+
10^{2}\operatorname{MSE}(D_{B_x},0)
+
10^{2}\operatorname{MSE}(D_{B_y},0).
\end{equation}

\paragraph{Diffusion objective.}
The diffusion model is trained with an EDM denoising loss on residuals. Given the operator prediction \(Y_{op}\) and DNS trajectory \(Y_{\mathrm{DNS}}\), the diffusion target is
\(\Delta Y = Y_{\mathrm{DNS}}-Y_{op}\). The denoising network is trained to recover this residual  using the EDM-weighted MSE loss,
\begin{equation}
\mathcal{L}_{\mathrm{EDM}}
=
\mathbb{E}_{\sigma,\epsilon}
\left[
\lambda(\sigma)
\left\|
D_\theta(\Delta Y+\sigma\epsilon,\sigma,c)-\Delta Y
\right\|_2^2
\right],
\end{equation}
where \(c=Y_{op}\) is the conditioning trajectory. For Helmholtz projection in the diffusion model, we first reconstruct the full field \(Y_{op}+\Delta Y_\theta\), before projecting the velocity and magnetic fields. We 
use paired normalization for each vector field, with one shared scale for
$(u_x,u_y)$ and one shared scale for $(B_x,B_y)$, so that the Helmholtz projection acts on
physically consistent vector components.

\subsection{Vorticity and current-density loss weights}
\label{app:vorticity_current_weights}
\Tab{tab:vorticity_current_weight_ablation} shows that explicit losses for vorticity and current-density substantially improves derived-field accuracy. Relative to $(\lambda_\omega,\lambda_J)=(0,0)$, the $(2,5)$ configuration performs best for the aggregate derived-field error and achieves the lowest current-density error while maintaining good primary-field accuracy.

\begin{table*}[t]
\centering
\caption{Field-wise and aggregate relative \(L_2\) errors for the vorticity- and current-loss weight ablations, evaluated using the single-regime (\(Re=Rm=1000\)) three-channel scOT \((u_x,u_y,A)\) with transfer learning. Here, \(\lambda_\omega\) and \(\lambda_J\) weight the relative \(L_2\) losses on vorticity and current density in \Eq{eq:phase_loss}, respectively.}
\label{tab:vorticity_current_weight_ablation}
\setlength{\tabcolsep}{4pt}
\renewcommand{\arraystretch}{1.20}
\small
\begin{tabular}{lcccccc}
\toprule
\multirow{2}{*}{$(\lambda_\omega,\lambda_J)$}
& \multicolumn{4}{c}{\textbf{Avg. rel. $L_2$}}
& \multicolumn{2}{c}{\textbf{Aggregate rel. $L_2$}} \\
\cmidrule(lr){2-5}\cmidrule(lr){6-7}
& $u_{\rm avg}$
& $B_{\rm avg}$
& $\boldsymbol{\omega}$
& $\mathbf{J}$
& $\mathbf{P}$
& $\mathbf{D}$ \\

\midrule
\multicolumn{7}{c}{$Re=Rm=1000$} \\
\midrule
$(0,0)$
& $0.036$ & $0.334$ & $0.156$ & $2.002$ & $0.185$ & $1.079$ \\
$(1,1)$
& $0.037$ & $0.297$ & $0.141$ & $1.270$ & $0.167$ & $0.705$ \\
$(2,5)$
& $0.041$ & $0.247$ & $0.149$ & $0.754$ & $0.144$ & $\mathbf{0.451}$ \\
$(5,2)$
& $0.033$ & $0.249$ & $0.106$ & $0.939$ & $\mathbf{0.141}$ & $0.522$ \\
$(5,5)$
& $0.064$ & $0.318$ & $0.183$ & $0.914$ & $0.191$ & $0.549$ \\
\bottomrule
\end{tabular}%
\end{table*}

\subsection{Training Details}
\label{app:training_details}
\subsubsection{Transfer Learning }
\label{app: transfer_learning}
\paragraph{POSEIDON-to-MHD initialization.}
We initialize the scOT backbone from the pretrained POSEIDON-T checkpoint
\citep{POSEIDON2024}. Since the pretrained model was trained on fluid-dynamics
operators, its input and output projections do not contain all MHD channels. We
therefore expand the patch-embedding and patch-recovery layers to the MHD channel
dimension. Parameters corresponding to the velocity channels are copied from the
pretrained model. Newly introduced magnetic-channel parameters are initialized
independently, while all compatible internal scOT encoder and decoder weights are
loaded from the pretrained checkpoint. This gives the model a pretrained
representation for fluid motion while leaving MHD-specific degrees of freedom
trainable.

For single-regime MHD fine-tuning, we use separate optimizer groups. The
pretrained scOT parameters use a smaller learning rate, while the expanded
input/output projection parameters use a larger learning rate. This split is
important because the velocity channels already have a useful initialization,
whereas the magnetic field channels must be learned from scratch. Unless otherwise
specified, all models are trained with AdamW. The exact learning rates, weight
decay, batch size, and training duration can be found in the configuration files from the PHASE repository \href{https://github.com/PHASE-MHD/PHASE}{[link]}. 

\paragraph{Magnetic-channel initialization.}
Because the pretrained POSEIDON checkpoint does not contain magnetic field
channels, the expanded MHD input and output projections require initialization
for the new magnetic field components. We found this initialization to be important:
naively random magnetic-channel weights can make early optimization unstable and
can cause the model to rely too strongly on the pretrained velocity pathway.

For the input projection, we initialize the magnetic-channel weights from the
mean of the pretrained velocity-channel weights. This gives the magnetic input
channels a scale and spatial filtering behavior similar to the pretrained fluid
channels, while not imposing a specific magnetic solution. For the output
projection, we initialize the new magnetic-channel weights to zero. Thus, before
MHD fine-tuning, the model preserves the pretrained velocity prediction and
does not introduce arbitrary magnetic outputs. The magnetic field is then learned
from the MHD data and physics-informed losses during fine-tuning.

This asymmetric initialization reflects the structure of the transfer problem:
velocity evolution benefits from the pretrained fluid-dynamics operator, while
magnetic evolution is an added MHD-specific degree of freedom. Initializing the
magnetic input pathway from the velocity-channel statistics gives the new
magnetic channels a compatible representation scale, whereas zero-initializing
the magnetic output pathway avoids spurious magnetic predictions at the start of
training.
\paragraph{Optimization choices for transferred and new parameters.}
We use different learning rates for pretrained and newly introduced parameters.
The pretrained scOT encoder--decoder and velocity-channel projections are
updated with a smaller learning rate to preserve the useful fluid-dynamics
representation inherited from POSEIDON. The newly introduced magnetic-channel
parameters, output projection parameters, and \(Re/Rm\)-conditioning adapters are
trained with a larger learning rate because they must be learned from MHD data.

We also use channel-weighted losses because the magnetic field has a different
physical scale from the velocity field in our data. Without this weighting, the
optimization can be dominated by the easier velocity channels, delaying or
suppressing magnetic field learning. In the SR and MR
scOT experiments, magnetic field terms are therefore upweighted in the data and
physics-informed losses. This weighting is applied after the fields are placed
on physically consistent normalization scales.
\subsubsection{Physical Parameter Conditioning via Residual Adapters}
\label{subsec: param_conditioning}
\paragraph{Multi-regime Data.}
For multi-regime training, each batch is balanced across the sampled Reynolds
numbers. The model receives the Reynolds number and magnetic Reynolds number as
metadata in addition to the input fields and lead time. We use log-scaled
conditioning variables because the sampled Reynolds numbers span more than one
order of magnitude:
\[
    \tilde{r} =
    \frac{\log_{10} Re-\mu_{\log Re}}{\sigma_{\log Re}},
    \qquad
    \tilde{m} =
    \frac{\log_{10} Rm-\mu_{\log Re}}{\sigma_{\log Re}}.
\]
For the Reynolds numbers used in the turbulence experiments,
\(\{80,200,400,650,1000,1500,2050,2750,3600,4500\}\), these statistics are
\(\mu_{\log Re}=2.9756\) and \(\sigma_{\log Re}=0.5417\).

\paragraph{Output \& Transformer-Block Conditioners.}The final output FiLM conditioner is a two-layer MLP that maps
\((\tilde{r},\tilde{m})\) to channel-wise scale and shift parameters. Its final
linear layer is zero-initialized, so the modulation is initially an identity
residual correction. The deep adapters use the same conditioning variables, but
each scOT block has its own adapter and its own conditioning MLP. Each adapter
has the form
\[
    E_\ell(h) =
    W_{\ell}^{\mathrm{up}}
    \mathrm{GELU}
    \left(
    W_{\ell}^{\mathrm{down}}\mathrm{LN}(h)
    \right),
\]
with bottleneck dimension \(64\) in our main multi-regime experiments. The
conditioner produces a channel-wise gate \(g_\ell(\tilde{r},\tilde{m})\), and
the block output is updated as
\[
    h_\ell \leftarrow h_\ell + g_\ell(\tilde{r},\tilde{m})\odot E_\ell(h_\ell).
\]
The adapter up-projection and the final layer of the gate network are
zero-initialized. Thus, when training begins from the SR checkpoint,
the MR model initially implements the original SR operator,
and the adapters learn parameter-dependent corrections during fine-tuning.

\paragraph{Warm start for multi-regime training.}
The MR model is warm-started from the trained SR
\(Re=1000\) MHD scOT model. We choose this checkpoint because \(Re=1000\) lies
near the center of the sampled parameter range and already contains learned
velocity--magnetic field coupling. During MR fine-tuning, the inherited scOT
weights are trained with a lower learning rate than the newly introduced
conditioning adapters and output FiLM layers. This lets the model preserve the
accurate single-regime solution operator while learning smooth
parameter-dependent corrections across the Reynolds-number range.

\subsection{Turbulence: evaluation metrics and results}
\label{app:turb_results_and_eval}
\subsubsection{Field structure and time evolution}
\label{sec:turb_relL2}
\begin{figure}
    \centering
    \includegraphics[width=0.88\linewidth,
        clip]{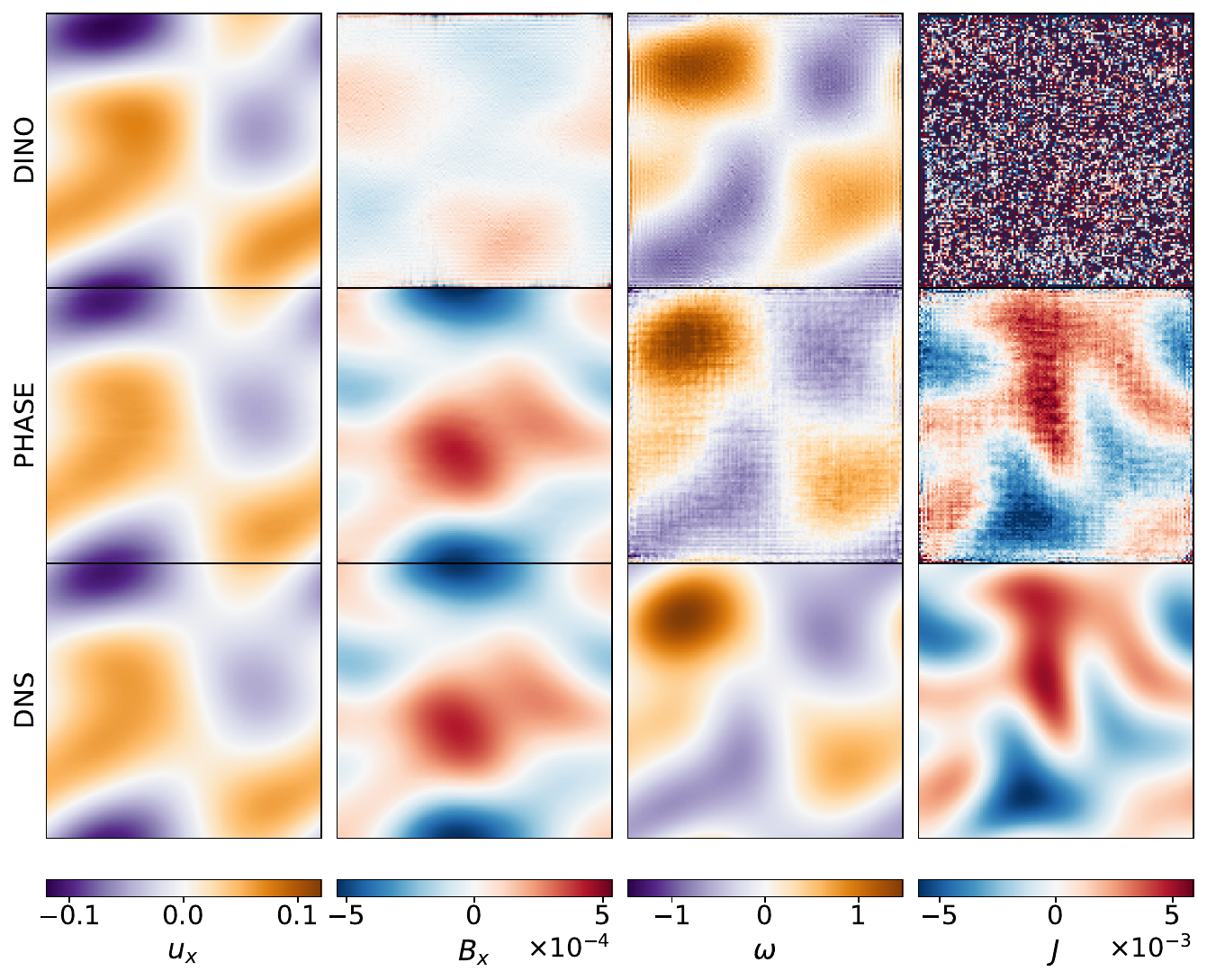}
    \caption{Same as \Fig{fig:fields_Re1000}, but for $Re=80$. PHASE outperforms DINO in the viscous regime, particularly in predicting the magnetic field and current-density.}
    \label{fig:fields_Re80}
\end{figure}

\begin{figure}
    \centering
    \includegraphics[width=0.88\linewidth,
        clip]{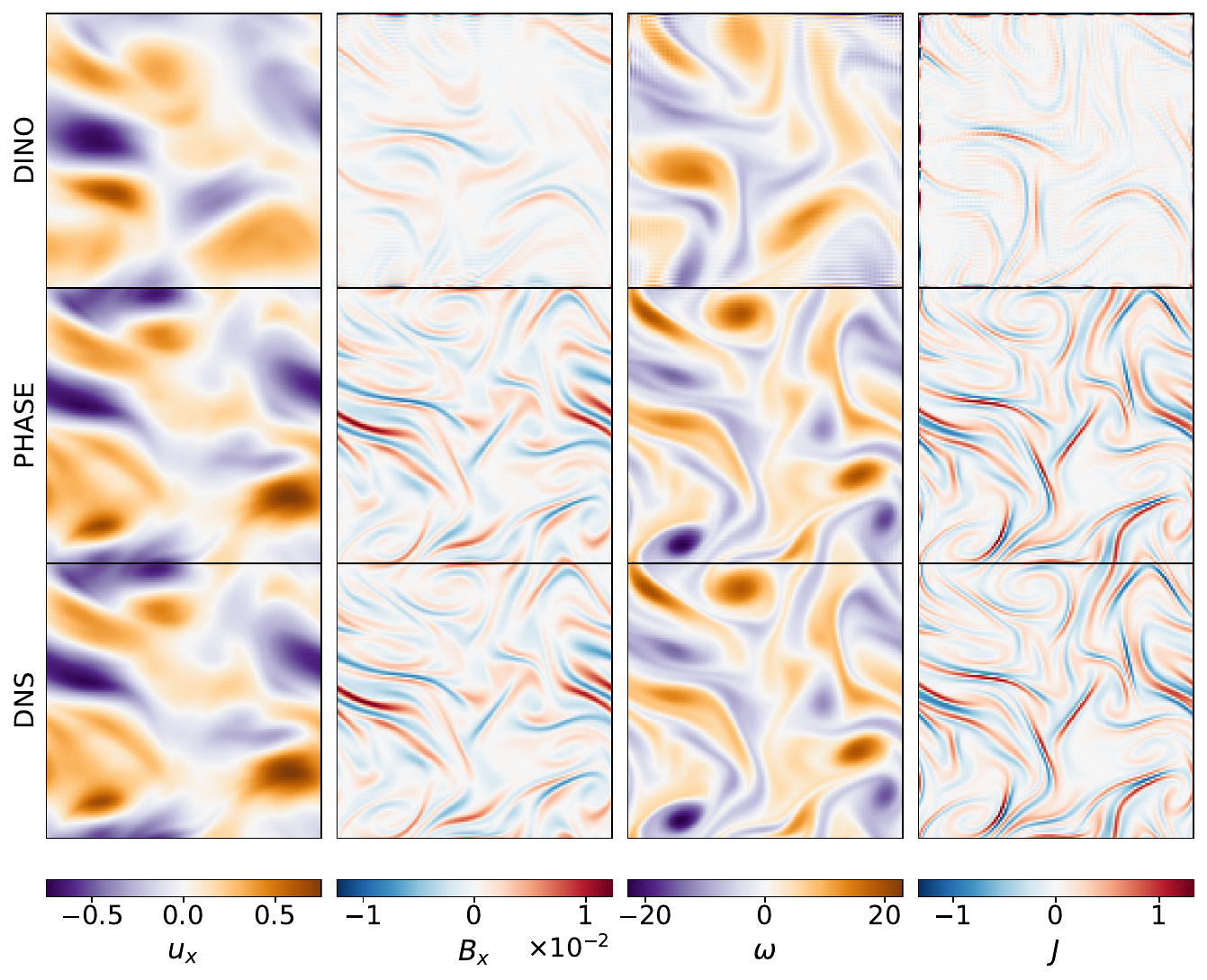}
    \caption{Same as \Fig{fig:fields_Re1000}, but for $Re=4500$. In this strongly turbulent regime, PHASE reproduces all fields more accurately than DINO.}
    \label{fig:fields_Re4500}
\end{figure}

\begin{figure}
    \centering
    \includegraphics[width=0.88\linewidth,
        clip]{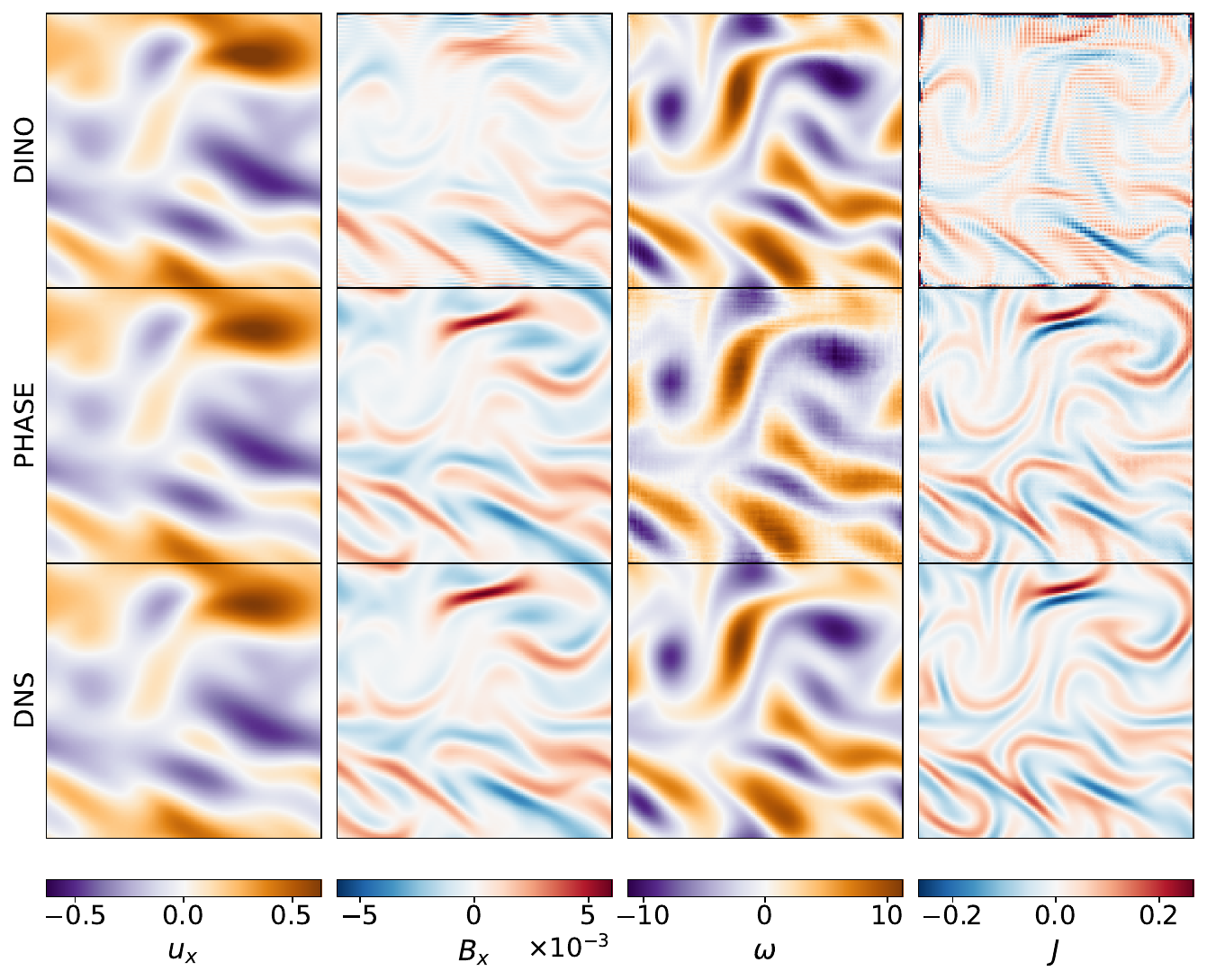}
    \caption{Same as \Fig{fig:fields_Re1000}, but for the unseen $Re=800$ generalization test. We compare DINO trained at $Re=1000$ with multi-regime PHASE. Although DINO captures the velocity and vorticity fields, it fails to recover the magnetic field and current-density for an unseen Reynolds number, whereas multi-regime PHASE accurately predicts all four fields.}
    \label{fig:fields_Re800}
\end{figure}

The field-wise and aggregate relative $L_2$ errors across the Reynolds-number regimes studied in this work, $Re=Rm=80$, $1000$, and $4500$, together with the unseen-$Re$ generalization test at $Re=800$, are reported in \Tab{tab:primary_derived_by_re_detailed}. All metrics are evaluated at each saved output time, averaged first over time within each sample, and then across all test samples. Across all regimes, both single-regime (SR) and multi-regime (MR) PHASE outperform DINO, with particularly large improvements for the magnetic and derived fields. PHASE remains accurate in both the highly viscous $Re=80$ and strongly turbulent $Re=4500$ regimes. 
Multi-regime training is particularly advantageous for generalization to unseen Reynolds numbers, with MR PHASE achieving the lowest primary-field and derived-field errors in the $Re=800$ generalization test. \Fig{fig:fields_Re80}, \Fig{fig:fields_Re4500}, and \Fig{fig:fields_Re800} compare the DINO and MR PHASE predictions with the DNS ground truth at the final simulation time, $t=1$, for $Re=80$, $Re=4500$, and $Re=800$, respectively.

\begin{table*}[t]
\centering
\caption{Field-wise and aggregate relative $L_2$ errors across Reynolds number regimes.}
\label{tab:primary_derived_by_re_detailed}
\setlength{\tabcolsep}{4pt}
\renewcommand{\arraystretch}{1.20}
\small
\begin{tabular}{lcccccc}
\toprule
\multirow{2}{*}{\textbf{Model}}
& \multicolumn{4}{c}{\textbf{Avg. rel. $L_2$}}
& \multicolumn{2}{c}{\textbf{Aggregate rel. $L_2$}} \\
\cmidrule(lr){2-5}\cmidrule(lr){6-7}
& $u_{\rm avg}$
& $B_{\rm avg}$
& $\boldsymbol{\omega}$
& $\mathbf{J}$
& $\mathbf{P}$
& $\mathbf{D}$ \\

\midrule
\multicolumn{7}{c}{$Re=Rm=1000$} \\
\midrule
DINO
& $0.035$ & $0.530$ & $0.081$ & $1.561$ & $0.283$ & $0.821$ \\
SR PHASE
& $0.029$ & $0.105$ & $0.069$ & $0.181$ & $0.067$ & $0.125$ \\
\textbf{MR PHASE}
& $\mathbf{0.012}$ & $\mathbf{0.044}$ & $\mathbf{0.063}$
& $\mathbf{0.108}$ & $\mathbf{0.028}$ & $\mathbf{0.085}$ \\

\midrule
\multicolumn{7}{c}{$Re=Rm=800$ (generalization test with unseen $Re$)} \\
\midrule
DINO
& $0.056$ & $0.518$ & $0.096$ & $1.715$ & $0.287$ & $0.905$ \\
SR PHASE
& $0.057$ & $0.172$ & $0.101$ & $0.312$ & $0.114$ & $0.206$ \\
\textbf{MR PHASE}
& $\mathbf{0.020}$ & $\mathbf{0.064}$ & $\mathbf{0.069}$
& $\mathbf{0.127}$ & $\mathbf{0.042}$ & $\mathbf{0.098}$ \\

\midrule
\multicolumn{7}{c}{$Re=Rm=80$ (viscous regime)} \\
\midrule
DINO
& $0.379$ & $1.171$ & $0.401$ & $6.470$ & $0.775$ & $3.435$ \\
\textbf{MR PHASE}
& $\mathbf{0.017}$ & $\mathbf{0.021}$ & $\mathbf{0.127}$
& $\mathbf{0.164}$ & $\mathbf{0.019}$ & $\mathbf{0.146}$ \\

\midrule
\multicolumn{7}{c}{$Re=Rm=4500$ (strong turbulence)} \\
\midrule
DINO
& $0.340$ & $0.824$ & $0.507$ & $1.506$ & $0.582$ & $1.007$ \\
\textbf{MR PHASE}
& $\mathbf{0.008}$ & $\mathbf{0.057}$ & $\mathbf{0.033}$
& $\mathbf{0.118}$ & $\mathbf{0.033}$ & $\mathbf{0.075}$ \\
\bottomrule
\end{tabular}%
\end{table*}

\begin{figure}
    \centering
    \includegraphics[width=\linewidth,
        clip]{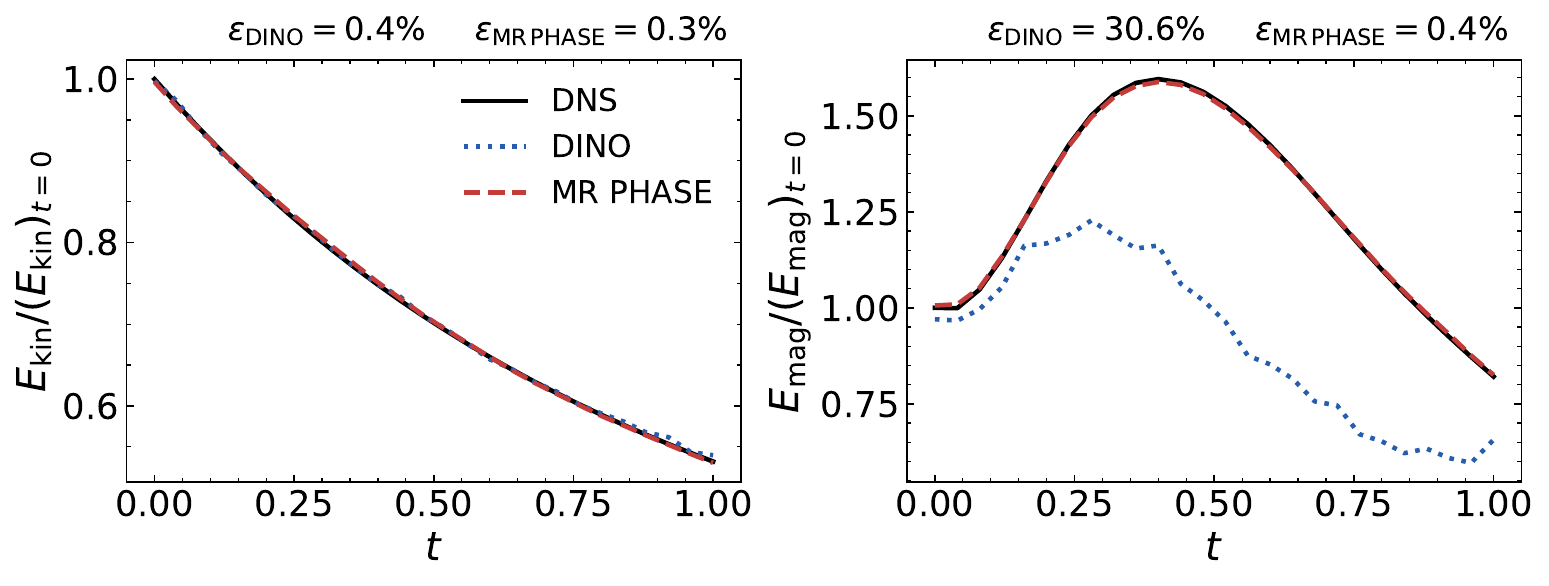}
    \caption{Time evolution of the normalized kinetic and magnetic energies for an $Re=Rm=1000$ test trajectory. Predictions from DINO and multi-regime PHASE are compared with the DNS, with each energy normalized by its corresponding DNS value at $t=0$. DINO accurately captures the kinetic-energy evolution but fails to reproduce the magnetic energy evolution, whereas multi-regime PHASE closely follows the DNS for both quantities.}
    \label{fig:turb_time_evol}
\end{figure}

In addition to the field-level comparisons, \Fig{fig:turb_time_evol} compares the time evolution of the kinetic and magnetic energies predicted by DINO and MR PHASE with the DNS ground truth. Both models reproduce the kinetic-energy decay, although MR PHASE follows the DNS more closely, with a relative root-mean-square (rms) error of $\varepsilon = 0.3\%$, compared with $0.4\%$ for DINO. The difference is more pronounced for the magnetic energy where MR PHASE accurately captures its initial growth and subsequent decay with a discrepancy of $0.4\%$, whereas DINO under predicts the magnetic energy and produces a substantially larger error of $31\%$. 

\subsubsection{Spectra and Probability density functions}
\label{sec:turb_spectra_pdfs}
Beyond point-wise field errors, we evaluate spectra and probability density functions (PDFs) to assess whether the models reproduce the multi-scale behavior and statistical structure of MHD turbulence. 

\begin{figure}
    \centering
    \includegraphics[width=1.0\linewidth,
        clip]{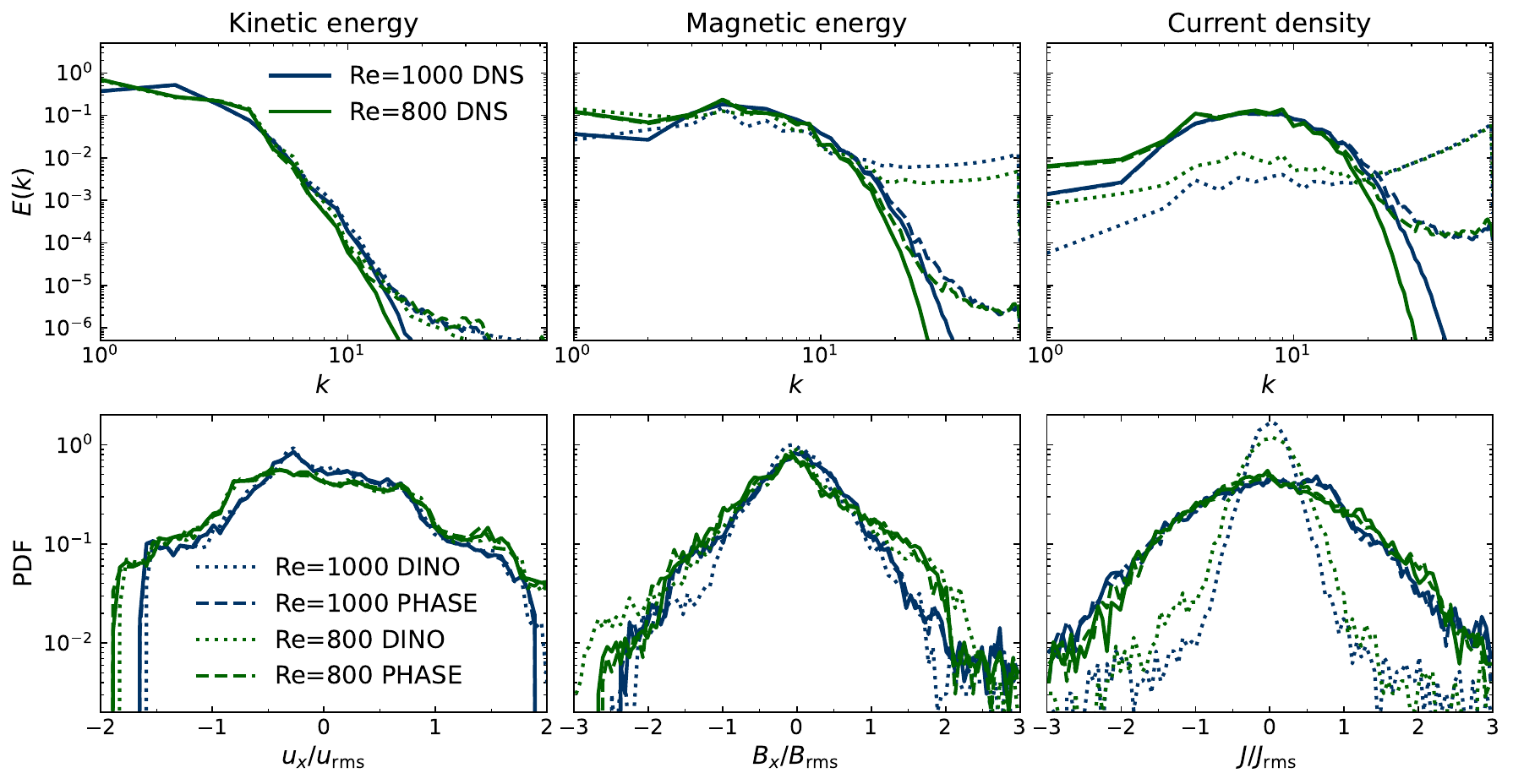}
    \caption{Same as \Fig{fig:spectra_pdfs_Re80_Re4500}, but for \(Re=Rm=1000\) and unseen \(Re=Rm=800\) tests. DINO and multi-regime PHASE predictions are compared with DNS. Compared to the DINO model, PHASE shows good predictions for all spectra and PDFs for both Reynolds numbers.}
    \label{fig:spectra_Re1000_Re800}
\end{figure}

\paragraph{Spectra}
Spectra quantify how kinetic and magnetic energy is distributed across spatial scales and are therefore sensitive to small-scale structures that may not be evident from aggregate field errors. For a vector field $\mathbf{q}=(q_x,q_y)$, we define the Fourier-space power as
\begin{equation}
P_{\mathbf{q}}(\mathbf{k})
=
|\widehat{q}_x(\mathbf{k})|^2+
|\widehat{q}_y(\mathbf{k})|^2,
\end{equation}
whereas for a scalar field in 2-D MHD, such as vorticity or current-density,
$P_q(\mathbf{k})=|\widehat{q}(\mathbf{k})|^2$. The spectrum is the
power summed over each integer Fourier shell,
\begin{equation}
E_q(k)
=
\sum_{\left||\mathbf{k}|-k\right|<0.5}P_q(\mathbf{k}).
\end{equation}
For the $128^2$ simulations, $k=1,\ldots,64$. We measure the spectral error at each shell in logarithmic space,
\begin{equation}
e(k)=
\left|
\log_{10}
\left(
\frac{E_{\mathrm{pred}}(k)}
     {E_{\mathrm{DNS}}(k)}
\right)
\right|.
\end{equation}
The low-wavenumber error is the mean of $e(k)$ over $k=1-8$, whereas the high-wavenumber error is its mean over $k=9-64$. Low-$k$ modes represent the large-scale flow, while high-$k$ modes probe small-scale velocity and magnetic structures. The low- and high-wavenumber errors for the kinetic-energy, magnetic energy, vorticity, and current-density spectra are reported across Reynolds-number regimes in \Tab{tab:spectrum_errors_by_re}. MR PHASE consistently achieves the lowest low-\(k\) errors for both primary and derived fields, with particularly accurate predictions in the strongly turbulent regime. At high \(k\), however, MR PHASE over predicts spectral power, especially at lower Reynolds numbers; a similar tendency is observed for DINO. Improving the recovery of this high-wavenumber spectral tail remains an important direction for future work. The spectra for $Re=1000$ and 800 are shown in \Fig{fig:spectra_Re1000_Re800}.

\begin{table*}[t]
\centering
\caption{Low- and high-wavenumber spectrum errors across Reynolds number regimes.}
\label{tab:spectrum_errors_by_re}
\setlength{\tabcolsep}{3.5pt}
\renewcommand{\arraystretch}{1.25}
\small
\begin{tabular}{lcccccccc}
\toprule
\multirow{2}{*}{\textbf{Model}}
& \multicolumn{4}{c}{\textbf{Low-$k$ spectrum error}}
& \multicolumn{4}{c}{\textbf{High-$k$ spectrum error}} \\
\cmidrule(lr){2-5}
\cmidrule(lr){6-9}
& $\mathbf{u}$
& $\mathbf{B}$
& $\boldsymbol{\omega}$
& $\mathbf{J}$
& $\mathbf{u}$
& $\mathbf{B}$
& $\boldsymbol{\omega}$
& $\mathbf{J}$ \\

\midrule
\multicolumn{9}{c}{$Re=Rm=1000$} \\
\midrule
DINO
& $0.018$ & $0.176$
& $0.018$ & $0.178$
& $4.573$ & $4.154$
& $\mathbf{4.500}$ & $4.155$ \\
SR PHASE
& $0.009$ & $0.013$
& $0.009$ & $0.013$
& $\mathbf{4.560}$ & $3.136$
& $4.659$ & $3.165$ \\
\textbf{MR PHASE}
& $\mathbf{0.005}$ & $\mathbf{0.005}$
& $\mathbf{0.005}$ & $\mathbf{0.005}$
& $4.686$ & $\mathbf{3.025}$
& $4.799$ & $\mathbf{3.051}$ \\

\midrule
\multicolumn{9}{c}{$Re=Rm=800$ (generalization test with unseen $Re$)} \\
\midrule
DINO
& $0.070$ & $0.130$
& $0.070$ & $0.131$
& $5.037$ & $5.194$
& $\mathbf{4.996}$ & $5.196$ \\
SR PHASE
& $0.073$ & $0.053$
& $0.073$ & $0.054$
& $\mathbf{4.992}$ & $3.958$
& $5.115$ & $4.012$ \\
\textbf{MR PHASE}
& $\mathbf{0.008}$ & $\mathbf{0.009}$
& $\mathbf{0.008}$ & $\mathbf{0.009}$
& $5.143$ & $\mathbf{3.663}$
& $5.285$ & $\mathbf{3.720}$ \\

\midrule
\multicolumn{9}{c}{$Re=Rm=80$ (viscous regime)} \\
\midrule
DINO
& $1.292$ & $1.236$
& $1.235$ & $1.240$
& $8.142$ & $8.767$
& $8.073$ & $8.769$ \\
\textbf{MR PHASE}
& $\mathbf{1.069}$ & $\mathbf{0.454}$
& $\mathbf{1.077}$ & $\mathbf{0.459}$
& $\mathbf{7.714}$ & $\mathbf{7.477}$
& $\mathbf{8.001}$ & $\mathbf{7.747}$ \\

\midrule
\multicolumn{9}{c}{$Re=Rm=4500$ (strong turbulence)} \\
\midrule
DINO
& $0.150$ & $0.378$
& $0.157$ & $0.381$
& $2.965$ & $1.952$
& $2.733$ & $1.952$ \\
\textbf{MR PHASE}
& $\mathbf{0.002}$ & $\mathbf{0.008}$
& $\mathbf{0.002}$ & $\mathbf{0.008}$
& $\mathbf{1.505}$ & $\mathbf{0.891}$
& $\mathbf{1.515}$ & $\mathbf{0.896}$ \\
\bottomrule
\end{tabular}%
\end{table*}

\paragraph{Probability density functions}
PDFs characterize the statistics of the predicted fields and provide a measure of distribution shape, fluctuation amplitude, and intermittency. At each time snapshot, the predicted and DNS fields are non-dimensionalized using the corresponding DNS rms:
\begin{align}
u_{\mathrm{rms}}
&=
\sqrt{\langle u_{x,\mathrm{DNS}}^2+u_{y,\mathrm{DNS}}^2\rangle},
&
B_{\mathrm{rms}}
&=
\sqrt{\langle B_{x,\mathrm{DNS}}^2+B_{y,\mathrm{DNS}}^2\rangle}, \\
\omega_{\mathrm{rms}}
&=
\sqrt{\langle\omega_{\mathrm{DNS}}^2\rangle},
&
J_{\mathrm{rms}}
&=
\sqrt{\langle J_{\mathrm{DNS}}^2\rangle}.
\end{align}

The PDF relative mean error is
\begin{equation}
\mathcal{E}_{\mathrm{PDF}}(q)
=
\frac{1}{|\mathcal{I}|}
\sum_{i\in\mathcal{I}}
\frac{|p_{\mathrm{pred},i}-p_{\mathrm{DNS},i}|}
     {p_{\mathrm{DNS},i}},
\qquad
\mathcal{I}=\{i:p_{\mathrm{DNS},i}>0\}.
\end{equation}
We use 60 bins over $[-2,2]$ for the velocity and magnetic field components and 90 bins over $[-3,3]$ for vorticity and current-density. We additionally evaluate the fluctuation amplitude using the relative standard-deviation error of the un-normalized physical fields,
\begin{equation}
\mathcal{E}_{\sigma}(q)
=
\frac{|\sigma(q_{\mathrm{pred}})-\sigma(q_{\mathrm{DNS}})|}
     {\sigma(q_{\mathrm{DNS}})}.
\end{equation}
Intermittency is studied using the kurtosis,
\begin{equation}
\kappa(q)
=
\left\langle
\left(
\frac{q-\langle q\rangle}{\sigma(q)}
\right)^4
\right\rangle,
\qquad
\mathcal{E}_{\kappa}(q)
=
|\kappa(q_{\mathrm{pred}})-\kappa(q_{\mathrm{DNS}})|.
\end{equation}
The kurtosis error is sensitive to sharp vorticity and current-density structures. The PDF, relative standard-deviation, and  kurtosis errors are reported across Reynolds-number regimes in \Tab{tab:distribution_errors_by_re}. MR PHASE consistently achieves the lowest errors for both primary and derived fields, demonstrating improved recovery of field distributions, fluctuation amplitudes, and intermittent structures across viscous, unseen, and strongly turbulent regimes. The improvement is particularly pronounced for current-density, whose heavy-tailed distribution is poorly captured by DINO. The PDFs for $Re=1000$ and the unseen $Re=800$ case are shown in \Fig{fig:spectra_Re1000_Re800}.

\begin{table*}[t]
\centering
\caption{PDF relative mean error, relative standard-deviation error, and absolute kurtosis errors across Reynolds number regimes.}
\label{tab:distribution_errors_by_re}
\setlength{\tabcolsep}{2.5pt}
\renewcommand{\arraystretch}{1.20}
\scriptsize
\begin{tabular}{lcccccccccccc}
\toprule
\multirow{2}{*}{\textbf{Model}}
& \multicolumn{4}{c}{\textbf{PDF error}}
& \multicolumn{4}{c}{\textbf{Standard-deviation error}}
& \multicolumn{4}{c}{\textbf{Kurtosis error}} \\
\cmidrule(lr){2-5}
\cmidrule(lr){6-9}
\cmidrule(lr){10-13}
& $\mathbf{u}_x$ & $\mathbf{B}_x$ & $\boldsymbol{\omega}$ & $\mathbf{J}$
& $\mathbf{u}_x$ & $\mathbf{B}_x$ & $\boldsymbol{\omega}$ & $\mathbf{J}$
& $\mathbf{u}_x$ & $\mathbf{B}_x$ & $\boldsymbol{\omega}$ & $\mathbf{J}$ \\

\midrule
\multicolumn{13}{c}{$Re=Rm=1000$} \\
\midrule
DINO
& $0.072$ & $0.284$ & $0.136$ & $0.329$
& $0.004$ & $0.152$ & $0.005$ & $0.623$
& $0.021$ & $1.159$ & $0.046$ & $48.278$ \\
SR PHASE
& $0.062$ & $0.103$ & $\mathbf{0.116}$ & $0.155$
& $0.003$ & $0.010$ & $0.003$ & $0.014$
& $0.017$ & $0.159$ & $0.027$ & $0.213$ \\
\textbf{MR PHASE}
& $\mathbf{0.039}$ & $\mathbf{0.075}$ & $0.117$ & $\mathbf{0.144}$
& $\mathbf{0.001}$ & $\mathbf{0.002}$ & $\mathbf{0.002}$ & $\mathbf{0.011}$
& $\mathbf{0.006}$ & $\mathbf{0.058}$ & $\mathbf{0.013}$ & $\mathbf{0.155}$ \\

\midrule
\multicolumn{13}{c}{$Re=Rm=800$ (generalization test with unseen $Re$)} \\
\midrule
DINO
& $0.167$ & $0.233$ & $0.232$ & $0.282$
& $0.036$ & $0.093$ & $0.058$ & $0.832$
& $0.024$ & $0.890$ & $0.033$ & $44.610$ \\
SR PHASE
& $0.174$ & $0.180$ & $0.236$ & $0.447$
& $0.039$ & $0.083$ & $0.060$ & $0.186$
& $0.028$ & $0.235$ & $0.038$ & $0.325$ \\
\textbf{MR PHASE}
& $\mathbf{0.053}$ & $\mathbf{0.084}$ & $\mathbf{0.118}$ & $\mathbf{0.151}$
& $\mathbf{0.003}$ & $\mathbf{0.006}$ & $\mathbf{0.002}$ & $\mathbf{0.009}$
& $\mathbf{0.013}$ & $\mathbf{0.073}$ & $\mathbf{0.022}$ & $\mathbf{0.132}$ \\

\midrule
\multicolumn{13}{c}{$Re=Rm=80$ (viscous regime)} \\
\midrule
DINO
& $0.210$ & $0.771$ & $0.299$ & $0.863$
& $0.037$ & $0.421$ & $0.055$ & $5.368$
& $0.108$ & $2.682$ & $0.227$ & $55.397$ \\
\textbf{MR PHASE}
& $\mathbf{0.045}$ & $\mathbf{0.053}$ & $\mathbf{0.130}$ & $\mathbf{0.140}$
& $\mathbf{0.003}$ & $\mathbf{0.005}$ & $\mathbf{0.008}$ & $\mathbf{0.014}$
& $\mathbf{0.009}$ & $\mathbf{0.011}$ & $\mathbf{0.030}$ & $\mathbf{0.080}$ \\

\midrule
\multicolumn{13}{c}{$Re=Rm=4500$ (strong turbulence)} \\
\midrule
DINO
& $0.242$ & $0.464$ & $0.288$ & $0.518$
& $0.052$ & $0.329$ & $0.083$ & $0.456$
& $0.183$ & $2.580$ & $0.317$ & $47.557$ \\
\textbf{MR PHASE}
& $\mathbf{0.032}$ & $\mathbf{0.082}$ & $\mathbf{0.086}$ & $\mathbf{0.147}$
& $\mathbf{0.001}$ & $\mathbf{0.004}$ & $\mathbf{0.001}$ & $\mathbf{0.010}$
& $\mathbf{0.005}$ & $\mathbf{0.094}$ & $\mathbf{0.009}$ & $\mathbf{0.250}$ \\
\bottomrule
\end{tabular}%
\end{table*}

\subsection{Kelvin-Helmoltz instability : results}
\label{app:kh_results}
To evaluate PHASE beyond decaying turbulence, we test it on the MHD Kelvin--Helmholtz instability, which evolves from a weak transverse perturbation through linear growth, nonlinear vortex roll-up, mixing, and decay. Across both viscous ($Re=200$) and turbulent ($Re=2050$) regimes, PHASE reproduces the passive-tracer evolution and instability structures in close agreement with the DNS, demonstrating its ability to capture MHD instability dynamics.

\begin{figure}
    \centering
    \includegraphics[width=\linewidth,
        clip]{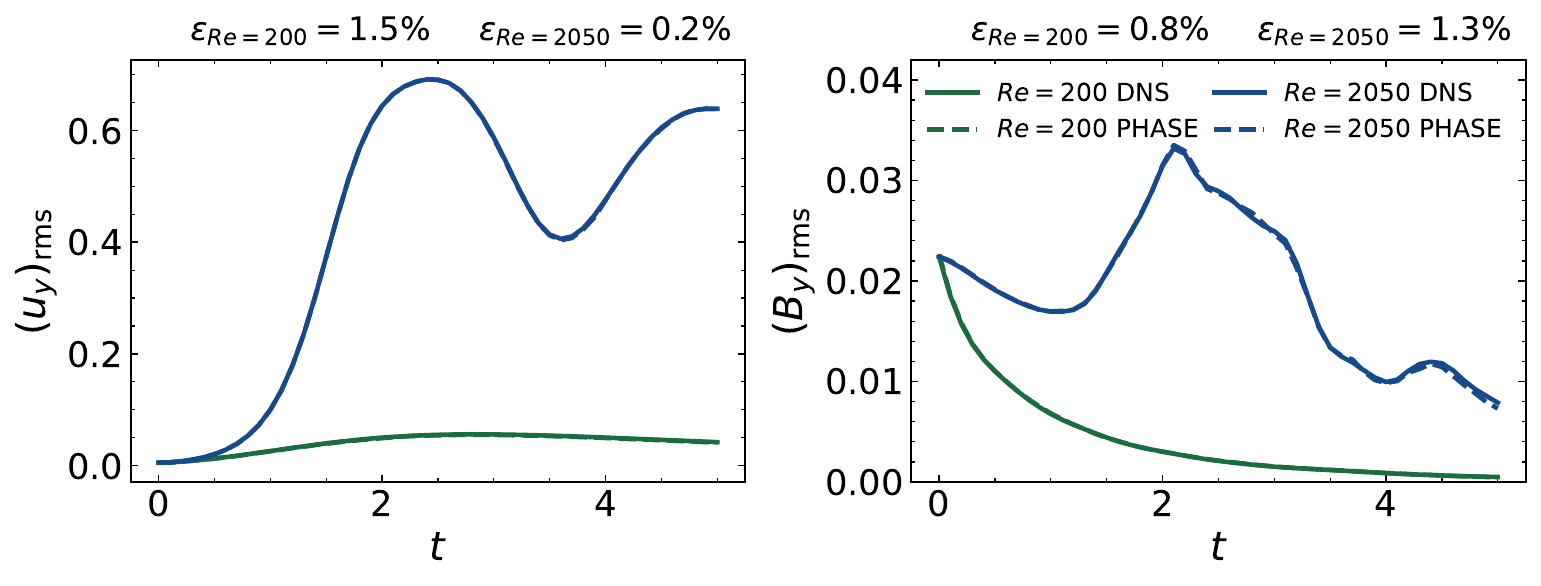}
    \caption{Time evolution of the root-mean-square (rms) value of the transverse velocity, \((u_y)_{\mathrm{rms}}\), and magnetic field, \((B_y)_{\mathrm{rms}}\), for a Kelvin--Helmholtz test trajectory at \(Re=Rm=200\) and \(2050\). Multi-regime PHASE predictions are compared with DNS over the entire simulation time. We also report  the relative temporal rms error, $\varepsilon$, for each Reynolds number.}
    \label{fig:kh_time_evol}
\end{figure}

\Fig{fig:kh_time_evol} compares the rms evolution of the transverse velocity and magnetic field for a test trajectory with \(Re=Rm=200\) and \(2050\). PHASE closely follows the DNS throughout the time evolution, capturing the growth, nonlinear evolution and subsequent decay of the instability across both Reynolds-number regimes. The relative temporal rms errors remain \(\lesssim 1.5\%\) for all cases, showing accurate recovery of the evolving field amplitudes in addition to the spatial structures.

\begin{figure}
    \centering
    \includegraphics[width=\linewidth,
        clip]{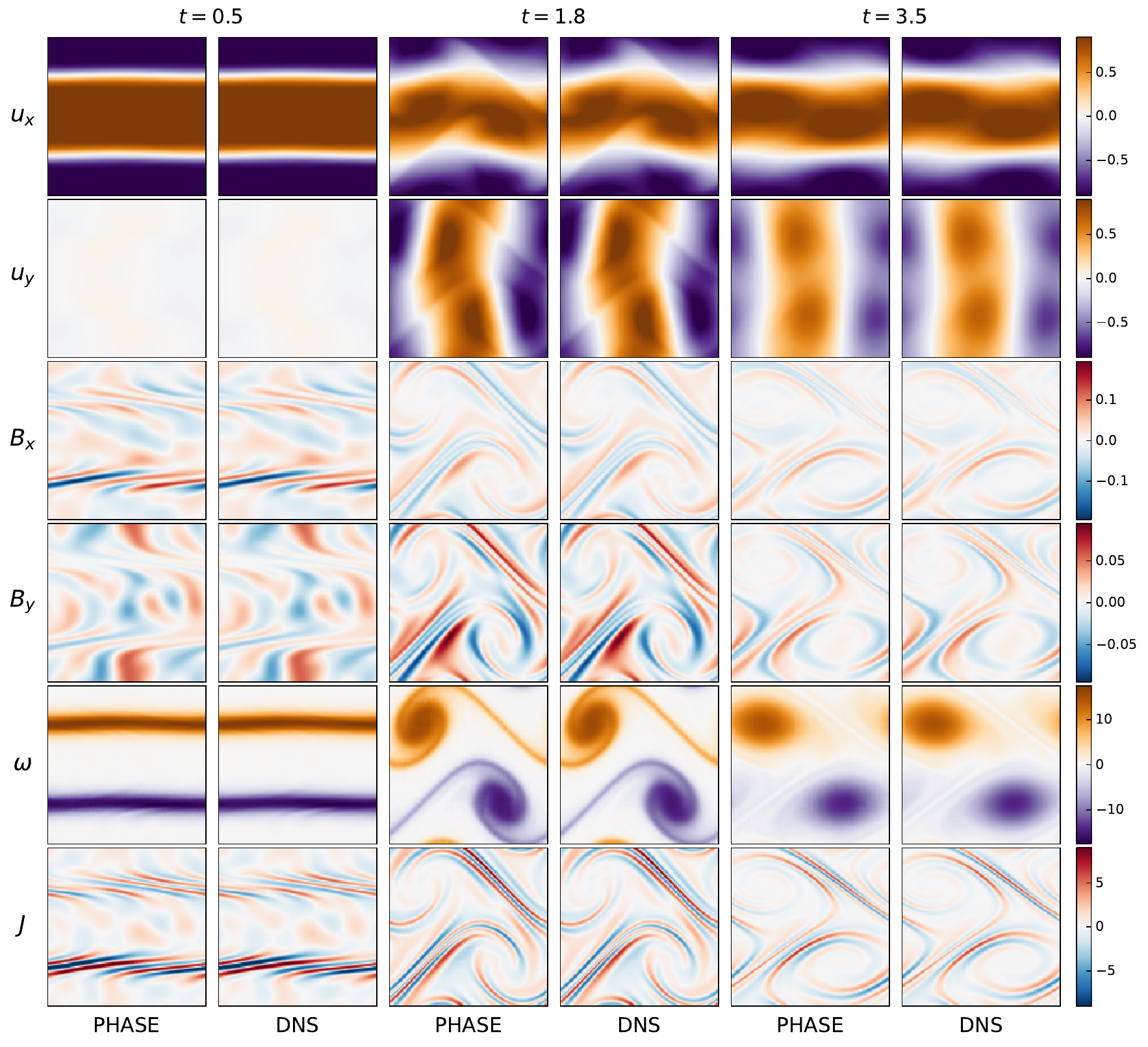}
    \caption{Comparison of multi-regime PHASE predictions and DNS for a Kelvin--Helmholtz test trajectory at \(Re=Rm=2050\). The velocity components, magnetic field components, vorticity, and current-density are shown at \(t=0.5\), \(1.8\), and \(3.5\), corresponding respectively to the initial, linear-growth, and nonlinear stages of the instability.}
    \label{fig:kh_fields_Re2050}
\end{figure}

\Fig{fig:kh_fields_Re2050} compares PHASE with DNS for a Kelvin--Helmholtz test sample at \(Re=Rm=2050\). PHASE accurately follows the evolution from the initial shear layers at \(t=0.5\), through linear instability growth at \(t=1.8\), to nonlinear roll-up at \(t=3.5\). The agreement extends beyond the primary velocity and magnetic fields to vorticity and current-density, demonstrating that the model recovers both the large-scale instability dynamics and the sharper structures in vorticity and current-density.

\end{document}